\documentclass[preprint,12pt,authoryear]{elsarticle}

\usepackage{amssymb}
\usepackage{amsmath}
\usepackage{amsthm}
\newtheorem{proposition}{Proposition}
\usepackage{graphicx}
\usepackage{booktabs}
\usepackage{multirow}
\usepackage{pgfplots}
\pgfplotsset{compat=1.17}

\usepackage{placeins}
\usepackage{url}
\begin{document}

\begin{frontmatter}

\title{SVD-Fusion CAM: Spatially Resolved Gradient Weighting and
Resolution-Aware Fusion for Class Activation Maps}

\author[1]{Newsha Shahbodaghkhan}
\ead{newsha.shah@aut.ac.ir}

\author[1]{Maryam Amirmazlaghani\corref{cor1}}
\ead{mazlaghani@aut.ac.ir}

\author[1]{Mohammad Rahmati}
\ead{rahmati@aut.ac.ir}

\author[2]{Mansoor Rezghi}
\ead{rezghi@modares.ac.ir}

\cortext[cor1]{Corresponding author}

\affiliation[1]{organization={Department of Computer Engineering, Amirkabir University of Technology},
            city={Tehran},
            country={Iran}}

\affiliation[2]{organization={Department of Computer Science, Tarbiat Modares University},
            city={Tehran},
            country={Iran}}

\begin{abstract}
Gradient-based Class Activation Mapping (CAM) methods such as Grad-CAM weight
each feature map by the spatial average of its gradient. We show that this
average is a projection onto a fixed spatial direction, the uniform vector,
and we prove that the resulting weights incur a larger first-order
attribution error than any rank-one estimate of the gradient matrix,
with equality only when the gradient is spatially flat. SVD-Fusion CAM
replaces the fixed direction with the dominant direction estimated by a
singular value decomposition of the joint spatial--channel gradient matrix,
and keeps the low-rank reconstruction as a spatial weight map rather than a
scalar. The map minimises the first-order attribution error among all
rank-$k$ weightings, contains Grad-CAM and HiResCAM as the two limiting
cases, and is invariant to the sign ambiguity of singular vectors. We further
treat the choice of explained layers as a design decision in its own right:
because a map cannot resolve detail finer than one feature-map cell, the
number of cells a target occupies bounds the attainable precision, which
yields a layer-selection criterion and predicts that multi-scale fusion
helps in proportion to how small the target is. Experiments on BRISC 2025
brain MRI and PASCAL VOC 2012 against eight baselines confirm this
prediction. On brain MRI the method improves localisation lift by $14.4\%$
and pointing-game accuracy by $71\%$ over HiResCAM, the strongest baseline,
with large paired effect sizes; on the larger objects of VOC it remains the
best method in lift and pointing game with a smaller margin. The method adds
one thin SVD to a single backward pass, a $3\%$ overhead over Grad-CAM.
\end{abstract}


\end{frontmatter}

\section{Introduction}
\label{sec:intro}

Deep convolutional networks now support decisions in domains where an incorrect
prediction carries real cost, from clinical triage to autonomous
navigation~\citep{tjoa2020survey, van2022explainable}. In such settings a
prediction is of limited value unless the evidence behind it can be inspected,
which has made post-hoc explanation methods a practical necessity rather than an
academic exercise~\citep{arrieta2020explainable}.

Among these methods, Class Activation Mapping (CAM) explains a prediction as a
weighted sum of the feature maps of a convolutional layer. The original
CAM~\citep{zhou2016learning} derives these weights from the classifier and
therefore applies only to networks that end in global average pooling, so other
architectures must be modified and retrained before they can be explained.
Gradient-based variants such as Grad-CAM~\citep{selvaraju2017grad} remove this
restriction: they require no architectural change, no retraining and only a
single backward pass. Because the explanation is a weighted sum, its quality
depends on how the weights are obtained. Grad-CAM and many of its descendants
obtain them by spatially averaging the gradient of the class score with respect
to each channel.

This averaging step rests on an implicit assumption that is rarely examined.
Averaging a gradient map into a scalar is usually presented as a summarisation
step, but it is in fact a projection onto a single fixed spatial direction, the
uniform vector (Sec.~\ref{sec:prelim}). The method therefore commits in advance
to the belief that a channel's relevance is the same everywhere in the feature
map. This belief is defensible when the object fills the frame and questionable
when it does not. A small lesion in a medical scan, for example, contributes
gradient energy in a narrow region, and averaging dilutes it into the surrounding
background. Our first observation is that this fixed direction need not be
assumed, because it can be estimated from the gradients themselves.

Replacing the uniform direction with the dominant singular direction of the
gradient matrix has three consequences. It yields a spatial weight map rather
than a scalar weight, so channels are modulated locally. It is optimal in a
precise sense: the first-order change of the class score under a perturbation
of the activations is an inner product with the gradient matrix, and we show
(Proposition~\ref{prop:attribution}) that the rank-$k$ reconstruction minimises
the error with which the explanation reproduces this change, whereas the
uniform projection of Grad-CAM can only do worse. And the resulting formulation
is not a new method disconnected from prior work but a relaxation of it:
Grad-CAM occupies the most constrained cell of a two-axis design space that the
decomposition makes explicit, and HiResCAM is its untruncated limit.

A second question, independent of how the weights are computed, is which layers
should be explained. Most CAM methods explain the final convolutional layer,
whose semantics are strongest and whose spatial grid is coarsest. Methods that
aggregate several layers, such as LayerCAM~\citep{jiang2021layercam}, improve on
this but do not specify how the layers should be chosen, treating the choice as
an implementation detail. We argue that the choice is not incidental. An
explanation computed at a given layer has one value per feature-map cell, so it
cannot resolve detail finer than a cell, and the number of cells that a target
occupies therefore limits the precision that any weighting scheme can achieve at
that layer. This observation gives an explicit selection criterion that depends
only on the feature-map geometry, and it leads to a testable prediction:
multi-scale fusion should help in proportion to how small the target is
relative to the image, and should bring little benefit when targets are large.
A geometric argument places the transition at about four cells.

We test this prediction on two datasets chosen for their contrast in target
size. On PASCAL VOC 2012 the annotated objects occupy a large fraction of the
image, whereas on the BRISC 2025 brain-tumour dataset the tumours occupy a few
percent. The observed ordering matches the prediction: the advantage over the
strongest baseline is larger for small targets than for large ones, between
$8.1\%$ and $20.6\%$ for targets covering less than a tenth of the image
against $4.5\%$ for targets covering more than a third of it, and the gain
contributed by fusion itself falls from $21$ pointing-game points for the
smallest objects to $1$ point for the largest.

The main contributions of this paper are as follows.
\begin{itemize}
  \item We show that scalar gradient aggregation is a projection onto a fixed
  spatial direction, and replace it with the dominant direction estimated by
  singular value decomposition. We prove that the resulting rank-$k$ spatial
  weighting minimises the first-order attribution error among all rank-$k$
  weightings and that Grad-CAM's error is never smaller than that of the
  rank-one estimate. The weighting is invariant to the sign ambiguity of
  singular vectors and contains Grad-CAM and HiResCAM as limiting cases of a
  two-axis taxonomy.
  \item We identify layer selection as a distinct and previously unspecified
  design decision, and give an explicit criterion based on how many feature-map
  cells the target occupies. The criterion depends only on the feature-map
  geometry, can be evaluated before any explanation is computed, and predicts
  when multi-scale fusion is worthwhile.
  \item We report a consistent relationship between target size and the benefit
  of multi-scale fusion across two datasets whose targets differ by more than an
  order of magnitude in relative size, together with the regime in which the
  benefit disappears.
  \item We evaluate the explanations along three complementary dimensions,
  namely localisation accuracy, faithfulness to the model and dependence on the
  learned parameters, under a protocol designed to be falsifiable: configuration
  selection on a held-out split, paired significance testing with effect sizes on
  every comparison, and per-class and per-size breakdowns.
\end{itemize}

The remainder of the paper is organised as follows. Sec.~\ref{sec:related}
reviews CAM variants and states where our approach differs from each family.
Sec.~\ref{sec:method} develops the method, and Sec.~\ref{sec:experiments}
reports the evaluation. Sec.~\ref{sec:practical} discusses practical
considerations for using the method, Sec.~\ref{sec:limitations} its scope,
and Sec.~\ref{sec:conclusion} concludes the paper.

\section{Related Work}
\label{sec:related}

We group prior work by the mechanism used to obtain feature-map weights, and
state in each case which limitation motivates our design. We then review two
issues that cut across all families: how layers are selected, and how saliency
maps are evaluated.

\subsection{Gradient-based Weighting}
CAM~\citep{zhou2016learning} projects the classifier weights onto globally
pooled feature maps and is restricted to architectures ending in global average
pooling. Grad-CAM~\citep{selvaraju2017grad} removes that restriction by
averaging gradients into per-channel scalars.
Grad-CAM++~\citep{chattopadhay2018gradcam} refines the averaging with
higher-order derivatives to handle multiple instances of a class, and
XGrad-CAM~\citep{fu2020axiom} derives weights that satisfy sensitivity and
conservation axioms. HiResCAM~\citep{draelos2020use} departs from the
scalar-per-channel view: rather than averaging the gradient, it multiplies
gradient and activation element-wise before summing over channels, which
preserves the per-location gradient sign and gives provable faithfulness
guarantees for certain architectures. It is a single-layer method and, in our
experiments (Sec.~\ref{sec:sota}), the strongest of the baselines we compare
against.

These methods differ in how the average is computed, but most of them agree
that the outcome should be a scalar per channel. The information lost in this
reduction is not noise: it is the spatial distribution of relevance within the
channel. Our method retains that distribution rather than refining the way it is
discarded. HiResCAM also keeps a spatial weighting, but from a single layer and
without separating the dominant gradient structure from the rest; our method
combines spatially resolved weights across layers of different resolution, which
is what produces the localisation gain on small targets.

\subsection{Gradient-free and Perturbation-based Weighting}
A second family avoids gradients altogether, motivated by saturation and noise
in backpropagated signals. Score-CAM~\citep{wang2020scorecam} weights each
channel by the change in class score when the input is masked by that channel's
activation, and Ablation-CAM~\citep{ramaswamy2020ablation} by the score drop when
the channel is removed. Both require one forward pass per channel, which is
prohibitive for wide layers. ReciproCAM~\citep{byun2022recipro} reduces this
cost by exploiting the correlation between spatially masked features and
outputs, and Opti-CAM~\citep{zhang2024opticam} optimises the combination weights
per image so that the logit of the masked image is maximised. On our setup
Score-CAM is roughly eight times slower than the gradient-based methods
(Sec.~\ref{sec:runtime}); we therefore restrict the main comparison to methods
of comparable cost.

These methods estimate relevance more directly than gradients do, but at a cost
that grows with the number of network evaluations. Our method keeps the
single-backward-pass cost of Grad-CAM, and its only additional expense is one
thin SVD of a matrix whose row count is the number of feature-map cells rather
than pixels.

\subsection{Decomposition-based Weighting}
Closest to our work are methods that apply matrix decompositions inside the CAM
pipeline. Eigen-CAM~\citep{muhammad2020eigen} projects the activations onto
their first principal component. SVD-Grad-CAM~\citep{thota2024svd} applies a
decomposition to each channel's two-dimensional gradient map, uses the low-rank
reconstruction as a denoiser, and then averages it into a scalar channel weight.
PCG-CAM~\citep{huang2025pcg} centres each channel's
gradient map, takes the leading eigenvector of the resulting covariance matrix,
projects the uncentred gradients onto it to obtain one weight per channel, and
takes the absolute value of the resulting map.
IP-CAM~\citep{zhang2025ipcam} also uses principal components, but of the
activations rather than the gradients, and it adds importance-weight layers to
the network, which must then be fine-tuned.
Tensor-decomposition variants extend this idea to higher-order factorisations of
the activation tensor~\citep{bouayed2024cnn}.

Two distinctions separate our approach from this family. First, decomposing
activations rather than gradients removes the dependence on the target class.
Our sanity checks (Sec.~\ref{sec:sanity}) show empirically that such maps barely
change when the classifier head is randomised, which makes them unsuitable as
class-discriminative explanations. Second, and more subtly, applying a
decomposition to gradients and then reducing it to one scalar weight per channel,
whether by averaging a per-channel reconstruction as in SVD-Grad-CAM or by
projecting the gradients onto a principal direction as in PCG-CAM, discards the
spatial structure the decomposition recovers. PCG-CAM also corroborates our sign
analysis. A singular or eigen vector is
defined only up to sign, so any scheme that reads a scalar off one must dispose
of the sign. PCG-CAM does so by taking a magnitude, and its own ablation shows
what happens otherwise: removing the absolute value drops its segmentation score
from $47.6$ to $10.2$ mIoU, a collapse of the same kind we measure in
Table~\ref{tab:taxonomy} for the unsigned scalar variant. The ambiguity does not
arise in our formulation. Sec.~\ref{sec:signs} shows that the outer product
$u_1v_1^{\top}$ is invariant to the joint sign flip, so no magnitude or sign
correction is needed. We decompose the joint spatial--channel gradient matrix
and keep the reconstruction as a spatial field rather than collapsing it.
Sec.~\ref{sec:ablation} isolates this difference in a controlled comparison in
which the two designs share the same backbone, the same layers and the same
fusion rule.

\subsection{Layer Selection and Multi-layer Aggregation}
Most methods explain a single layer, almost always the last convolutional one.
LayerCAM~\citep{jiang2021layercam} observes that shallower layers carry finer
spatial detail and reweights gradients element-wise so that earlier layers become
usable. Several other methods aggregate information from more than one layer.
FullGrad~\citep{srinivas2019full} sums gradient-based maps from all layers of the
network, Zoom-CAM~\citep{shi2020zoom} integrates importance maps over the
intermediate layers to recover small objects, and Poly-CAM~\citep{englebert2024poly}
combines earlier and later layers to obtain high-resolution maps. The most recent,
Winsor-CAM~\citep{wall2026winsor}, aggregates Grad-CAM maps from all
convolutional layers and suppresses outliers by Winsorising at a user-selected
percentile, reporting improved localisation when earlier layers are included.

These methods confirm that aggregating across scales helps, but none of them
provides a criterion for which layers to combine. The set is fixed by
convention, by depth index, or, as in FullGrad and Winsor-CAM, taken to be all
layers, with any remaining parameter left to the practitioner. Including every
layer adds computation and mixes in levels that are redundant or off-target.
Moreover, in standard backbones the spatial resolution changes only at a few
stride transitions, so a set chosen by depth may contain a single resolution
level, in which case fusion averages redundant maps. We instead relate the
choice of levels to the expected size of the target relative to the
feature-map grid. This indicates when a higher-resolution level is needed,
which agrees with the outcome of an exhaustive search on a held-out split
(Sec.~\ref{sec:ablation}), and additionally predicts the regime
in which aggregation becomes harmful, a regime we identify empirically in
Sec.~\ref{sec:size_law}. We therefore treat layer selection as a first-class
design decision (Sec.~\ref{sec:fusion}).

\subsection{Evaluation of Saliency Maps}
Ground-truth annotations exist for the location of a target, and we use them
here in the form of boxes for VOC and masks for BRISC. They do not exist for
attribution, because the evidence a model relies on is internal and need not
coincide with the annotated region. Localisation metrics therefore use
annotations as a proxy, and a classifier that exploits surrounding context will
penalise even a faithful explanation. Two complementary families of metrics are
used. Localisation metrics include the pointing game~\citep{zhang2018top}, its
energy-based variant~\citep{wang2020scorecam} and IoU, while faithfulness metrics
such as deletion and insertion curves~\citep{petsiuk2018rise} avoid the proxy
assumption entirely. Average drop and average
increase~\citep{chattopadhay2018gradcam} were widely adopted, but
\citet{zhang2024opticam} identify a flaw in both and propose a corrected
alternative.

Following this critique, we do not report average drop or average increase.
Instead, we report the energy-based pointing game, the pointing game and IoU for
localisation, and deletion and insertion for faithfulness. We treat the two
families as complementary rather than interchangeable. A map may be unfaithful
yet well localised if the model exploits context correlated with the target, and
conversely faithful yet poorly localised if the model genuinely relies on
context, a case we encounter and analyse in Sec.~\ref{sec:perclass}. This is
consistent with the observation of \citet{zhang2024opticam} that localisation
and classifier interpretability need not be aligned. Because perturbation-based
metrics construct inputs the model has never seen, they are themselves
imperfect, and we interpret them accordingly. We additionally report cascading
weight-randomisation sanity checks~\citep{adebayo2018sanity}, since a method may
score well on both families while being largely independent of the model it
purports to explain.

\section{Methodology}
\label{sec:method}

\begin{figure*}[t]
\centering
\includegraphics[width=\textwidth]{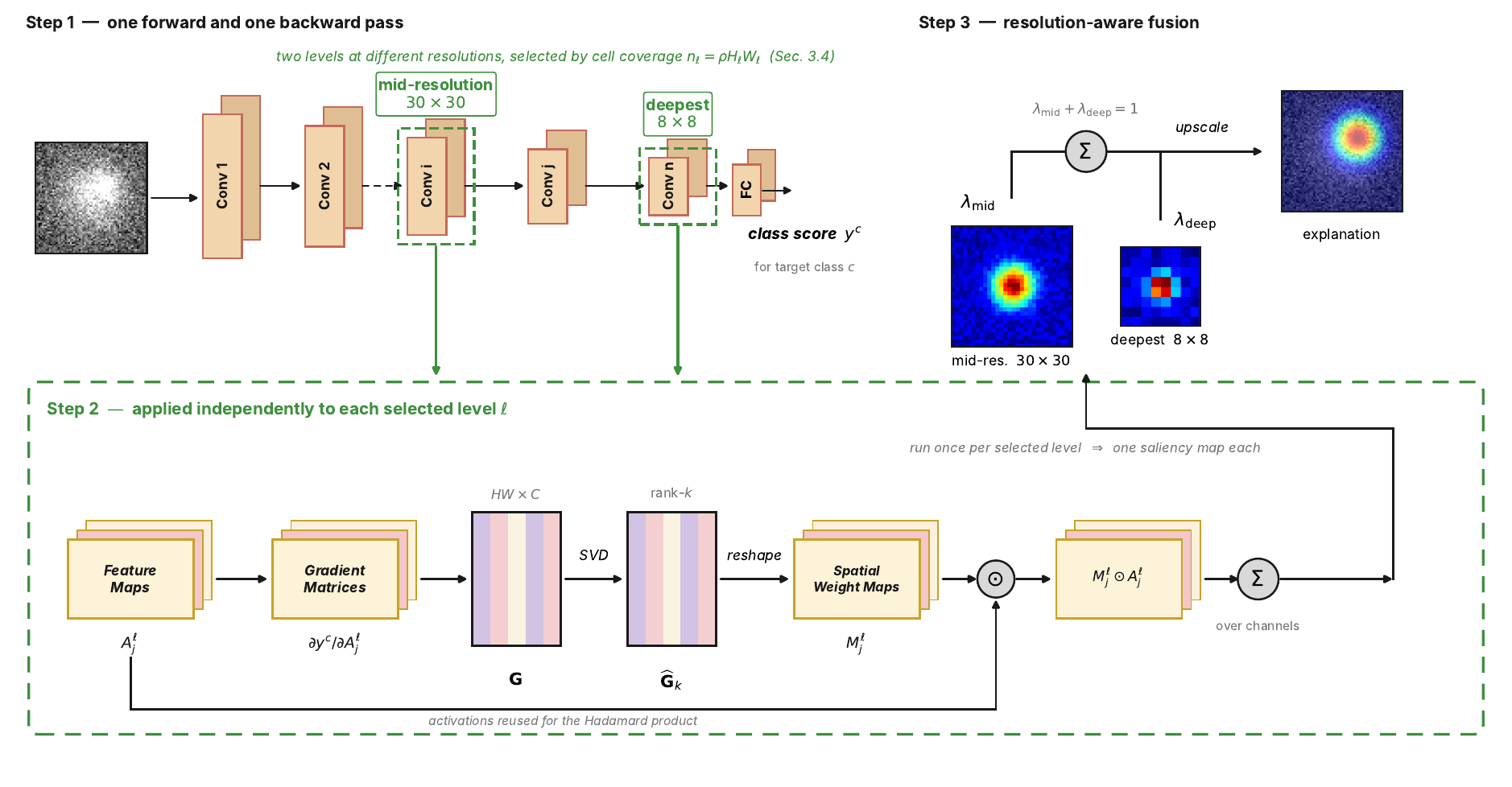}
\caption{Overview of SVD-Fusion CAM. \emph{Step 1}: layers are selected by
spatial resolution rather than by depth, so that the selected set spans distinct
grids; the criterion of Eq.~\eqref{eq:cells} determines which levels are needed.
\emph{Step 2}: for each selected layer the gradients of the target class score are
arranged into a single matrix whose rows index spatial positions and whose
columns index channels; its rank-$k$ reconstruction is reshaped into one spatial
weight map per channel and applied element-wise to the corresponding feature map.
\emph{Step 3}: the resulting maps are upsampled and combined. The resolution of
the selected levels bounds the precision attainable by the explanation. Symbols
follow Sec.~\ref{sec:method}: $y^{c}$ is the score of the target class,
$A^{\ell}_{j}$ the $j$-th activation channel of selected layer $\ell$,
$\mathbf{G}$ the gradient matrix of Eq.~\eqref{eq:gradmat}, $\widehat{\mathbf{G}}_k$
its rank-$k$ reconstruction, and $M^{\ell}_{j}$ the spatial weight map obtained by
reshaping column $j$.}
\label{fig:architecture}
\end{figure*}

We first show that existing gradient-based CAMs implicitly commit to a fixed
spatial profile when they reduce gradients to scalars (Sec.~\ref{sec:prelim}).
We then replace that fixed choice with the dominant profile estimated from the
gradients themselves, prove that this choice minimises the first-order
attribution error (Sec.~\ref{sec:spectral}--\ref{sec:svdk}), and finally
introduce a resolution-aware fusion rule together with a criterion that predicts
when fusion is needed (Sec.~\ref{sec:fusion}). Fig.~\ref{fig:architecture} gives
an overview of the three steps.

\subsection{Notation and the Limitation of Scalar Aggregation}
\label{sec:prelim}

Let $A^{l}\in\mathbb{R}^{H_l\times W_l\times C_l}$ be the activations of layer
$l$, where $H_l\times W_l$ is its spatial size and $C_l$ its number of channels.
We write $A^{l}_{j}$ for its $j$-th channel and $N_l=H_lW_l$ for its number of
spatial cells. When a single layer is considered, we drop the index $l$ and write
$A_j$, $N$ and $C$. For a target class $c$, let $y^{c}$ denote the class score at the output of the
classifier, which is the softmax probability for the single-label BRISC model and
the sigmoid probability for the multi-label VOC model; the same score is used for
all compared methods. We arrange all gradients of $y^{c}$ into a single gradient
matrix
\begin{equation}
\mathbf{G}^{c}\in\mathbb{R}^{N\times C},\qquad
\mathbf{G}^{c}[p,j]=\frac{\partial y^{c}}{\partial A_{j}(p)},
\label{eq:gradmat}
\end{equation}
where $p$ indexes spatial positions. Since the target class is fixed throughout,
we omit the superscript and write $\mathbf{G}$. Column $j$ of $\mathbf{G}$ is the
vectorised gradient map of channel $j$. We arrange the activations in the same
way as a matrix $\mathbf{A}\in\mathbb{R}^{N\times C}$.

Grad-CAM~\citep{selvaraju2017grad} assigns channel $j$ the scalar
$\alpha_j=\frac{1}{N}\sum_{p}\mathbf{G}[p,j]$, which in matrix form is
\begin{equation}
\boldsymbol{\alpha}=\tfrac{1}{N}\,\mathbf{G}^{\top}\mathbf{1},
\qquad \mathbf{1}=(1,\dots,1)^{\top}\in\mathbb{R}^{N}.
\label{eq:proj}
\end{equation}
Eq.~\eqref{eq:proj} states that Grad-CAM projects every gradient map onto one
fixed spatial direction, the uniform vector. This is equivalent to assuming that
the gradient signal has a flat spatial profile. When the evidence for a class is
concentrated in a sub-region, which is exactly the case for localised targets,
this fixed direction is a poor match and the resulting scalar discards the
information that localisation requires.

Our starting point is therefore simple: rather than fixing the spatial direction
in advance, we estimate it from the gradients.

\subsection{Spectral Gradient Weighting}
\label{sec:spectral}

To estimate the dominant spatial direction, we apply the thin singular value
decomposition~\citep{golub2013matrix}
\begin{equation}
\mathbf{G}=\mathbf{U}\boldsymbol{\Sigma}\mathbf{V}^{\top}
=\sum_{i=1}^{R}\sigma_i\,\mathbf{u}_i\mathbf{v}_i^{\top},
\qquad \sigma_1\ge\dots\ge\sigma_R>0,
\label{eq:svd}
\end{equation}
where $R=\operatorname{rank}(\mathbf{G})$. The vectors $\mathbf{u}_i\in\mathbb{R}^{N}$ are spatial modes and the
vectors $\mathbf{v}_i\in\mathbb{R}^{C}$ are channel modes. The pair
$(\mathbf{u}_1,\mathbf{v}_1)$ identifies the single spatial pattern and the
single channel combination that together explain the largest share of gradient
energy.

Keeping the leading $k$ terms,
$\widehat{\mathbf{G}}_{k}=\sum_{i\le k}\sigma_i\mathbf{u}_i\mathbf{v}_i^{\top}$,
is not a heuristic denoising step. By the Eckart--Young--Mirsky theorem it is
the optimal rank-$k$ approximation of $\mathbf{G}$,
\begin{equation}
\widehat{\mathbf{G}}_{k}
=\arg\min_{\operatorname{rank}(\mathbf{X})\le k}\lVert\mathbf{G}-\mathbf{X}\rVert_{F},
\label{eq:eckart}
\end{equation}
with retained energy fraction
$\eta_k=\sum_{i\le k}\sigma_i^{2}\big/\sum_{i\le R}\sigma_i^{2}$.
No other rank-$k$ matrix explains more of the gradient.

The optimality of Eq.~\eqref{eq:eckart} is a statement about the gradient
matrix, not yet about the explanation. The following proposition closes that
gap by relating the rank-$k$ weighting to the first-order behaviour of the
class score, which is what a faithful explanation is meant to capture.

\begin{proposition}[First-order attribution error]
\label{prop:attribution}
For any weight matrix $\mathbf{W}\in\mathbb{R}^{N\times C}$, define the
pre-ReLU map $L(\mathbf{W})=\sum_{j}\mathbf{W}[:,j]\odot\mathbf{A}[:,j]$, whose
total mass is $\mathbf{1}^{\top}L(\mathbf{W})=\langle\mathbf{W},\mathbf{A}\rangle_F$.
The first-order change of the class score under a perturbation
$\Delta\mathbf{A}$ of the activations is
$\Delta y^{c}\approx\langle\mathbf{G},\Delta\mathbf{A}\rangle_F$, and the map
built from $\mathbf{W}$ reproduces it with error
\begin{equation}
\big|\langle\mathbf{G},\Delta\mathbf{A}\rangle_F-\langle\mathbf{W},\Delta\mathbf{A}\rangle_F\big|
\;\le\;\lVert\mathbf{G}-\mathbf{W}\rVert_F\,\lVert\Delta\mathbf{A}\rVert_F .
\label{eq:attr_error}
\end{equation}
Among all $\mathbf{W}$ of rank at most $k$, the bound is minimised by
$\mathbf{W}=\widehat{\mathbf{G}}_k$, with value
$\sqrt{\sum_{i>k}\sigma_i^{2}}\;\lVert\Delta\mathbf{A}\rVert_F
=\sqrt{1-\eta_k}\,\lVert\mathbf{G}\rVert_F\,\lVert\Delta\mathbf{A}\rVert_F$.
Grad-CAM corresponds to
$\mathbf{W}_{\mathrm{GC}}=\tfrac{1}{N}\mathbf{1}\mathbf{1}^{\top}\mathbf{G}$,
a rank-one matrix whose spatial factor is fixed to $\mathbf{1}/\sqrt{N}$, and
satisfies
\begin{equation}
\lVert\mathbf{G}-\mathbf{W}_{\mathrm{GC}}\rVert_F
\;\ge\;\lVert\mathbf{G}-\widehat{\mathbf{G}}_1\rVert_F
=\sqrt{\textstyle\sum_{i\ge 2}\sigma_i^{2}},
\label{eq:gradcam_error}
\end{equation}
with equality if and only if $\mathbf{1}/\sqrt{N}$ is a leading left singular
vector of $\mathbf{G}$.
\end{proposition}

\begin{proof}
Eq.~\eqref{eq:attr_error} is the Cauchy--Schwarz inequality applied to
$\langle\mathbf{G}-\mathbf{W},\Delta\mathbf{A}\rangle_F$. The minimiser over
rank-$k$ matrices and the value of the minimum are given by the
Eckart--Young--Mirsky theorem, Eq.~\eqref{eq:eckart}. For
Eq.~\eqref{eq:gradcam_error}, write
$\mathbf{W}_{\mathrm{GC}}=\mathbf{p}\mathbf{p}^{\top}\mathbf{G}$ with
$\mathbf{p}=\mathbf{1}/\sqrt{N}$; this is the orthogonal projection of the
columns of $\mathbf{G}$ onto $\operatorname{span}\{\mathbf{p}\}$, hence a
rank-one matrix with residual
$\lVert\mathbf{G}-\mathbf{p}\mathbf{p}^{\top}\mathbf{G}\rVert_F^{2}
=\lVert\mathbf{G}\rVert_F^{2}-\lVert\mathbf{p}^{\top}\mathbf{G}\rVert_2^{2}$.
Over unit vectors $\mathbf{p}$, the term $\lVert\mathbf{p}^{\top}\mathbf{G}\rVert_2$
is at most $\sigma_1$, attained exactly by the leading left singular vectors,
so the residual is at least
$\lVert\mathbf{G}\rVert_F^{2}-\sigma_1^{2}=\sum_{i\ge2}\sigma_i^{2}$.
\end{proof}

Proposition~\ref{prop:attribution} makes precise the sense in which the
proposed weighting is faithful: of all explanations that modulate each channel
by a rank-$k$ spatial field, it is the one whose mass responds to activation
changes most nearly as the class score does, and the gap to the full gradient
is exactly the discarded spectral energy $1-\eta_k$. Grad-CAM is the special
case in which the spatial factor is not estimated but fixed, and
Eq.~\eqref{eq:gradcam_error} shows that this choice is never better and is
strictly worse whenever the dominant gradient mode is not spatially flat, which
is the situation of a localised target. The equality case is not hypothetical:
when the explained layer feeds global average pooling directly, as the last
block of a ResNet does, the gradient is identical at every position of each
channel, $\mathbf{u}_1=\mathbf{1}/\sqrt{N}$ exactly, and Grad-CAM, HiResCAM
and the proposed weighting coincide at that layer; the spatial information
must then come from a layer that is not followed immediately by the pooling,
which is also what makes the choice of layers in Sec.~\ref{sec:fusion} a
design decision in its own right. HiResCAM is the other limit,
$\mathbf{W}=\mathbf{G}$, with zero first-order error but no suppression of the
low-energy modes; Sec.~\ref{sec:ablation} shows empirically that the
intermediate choice localises better than either. The rank $k$ is therefore the
single parameter of the weighting. Measured on real images
(\ref{app:energy}), the gradient spectrum decays slowly: the leading mode
retains $30$--$35\%$ of the energy and the first five modes $63$--$69\%$. A
small fixed rank thus discards most of the gradient energy and yet localises
better than the full gradient, which shows that $k$ trades first-order
fidelity for spatial concentration rather than merely removing noise;
Sec.~\ref{sec:ablation} shows that the ranking of methods is the same for
every $k\in\{1,\dots,5\}$.

This use of the decomposition differs from that of earlier decomposition-based
methods in what is decomposed and in what is kept.
Eigen-CAM~\citep{muhammad2020eigen} decomposes the activations rather than the
gradients. Activations do not depend on the target class, so the resulting map
is class-agnostic, and our sanity checks (Sec.~\ref{sec:sanity}) confirm that
its output barely changes when the classifier head is randomised.
SVD-Grad-CAM~\citep{thota2024svd} does decompose gradients, but it treats each
channel's two-dimensional gradient map separately and then averages the
reconstruction into a scalar, so the decomposition acts purely as a denoiser
before the same scalar reduction as Grad-CAM is applied. We instead decompose
the joint spatial--channel matrix $\mathbf{G}\in\mathbb{R}^{N\times C}$ and keep
the reconstruction as a spatial field. Sec.~\ref{sec:ablation} isolates this
difference experimentally.

\subsection{From Channel Weights to Spatial Weight Maps}
\label{sec:svdk}

The decomposition can be turned into an explanation in two natural ways, which
we treat as two variants of one framework. The first, the \emph{channel-only}
variant, which we denote \textsc{svd1}, uses the principal channel mode
directly as channel weights,
\begin{equation}
L^{c}_{\textsc{svd1}}
=\mathrm{ReLU}\Big(\textstyle\sum_{j=1}^{C} v_1[j]\,A_{j}\Big).
\label{eq:svd1}
\end{equation}
This replaces the column means of Grad-CAM with the dominant channel direction,
but each channel is still scaled uniformly over space.

The second, the \emph{spatial} variant, is the one we propose and is what we
call SVD-Fusion CAM when computed at a single layer. Instead of reducing
column $j$ of $\widehat{\mathbf{G}}_{k}$ to a scalar, it reshapes this column
back to the spatial grid, which gives a weight map
$\mathbf{M}_{j}\in\mathbb{R}^{H\times W}$ for each channel, and computes
\begin{equation}
L^{c}_{k}
=\mathrm{ReLU}\Big(\textstyle\sum_{j=1}^{C}\mathbf{M}_{j}\odot A_{j}\Big),
\label{eq:svdk}
\end{equation}
where $\odot$ is the Hadamard product. Each channel is now modulated locally:
positions belonging to the dominant gradient mode are amplified and positions
outside it are suppressed. In the notation of
Proposition~\ref{prop:attribution}, Eq.~\eqref{eq:svdk} is
$\mathrm{ReLU}(L(\widehat{\mathbf{G}}_k))$.

The two variants are not competing designs; the second contains the first. For
$k=1$, column $j$ of $\widehat{\mathbf{G}}_{1}$ equals
$\sigma_1 v_1[j]\,\mathbf{u}_1$, which means that all columns are the same
spatial vector, rescaled per channel. Factoring this vector out of
Eq.~\eqref{eq:svdk} gives
\begin{equation}
L^{c}_{1}
=\mathrm{ReLU}\Big(\sigma_1\,\mathbf{U}_1\odot
\textstyle\sum_{j=1}^{C}v_1[j]\,A_{j}\Big),
\label{eq:factorised}
\end{equation}
with $\mathbf{U}_1$ the reshaping of $\mathbf{u}_1$ to the spatial grid.
Comparing Eq.~\eqref{eq:factorised} with Eq.~\eqref{eq:svd1} shows that, before
the ReLU, the spatial variant is exactly the channel-only variant gated by the
estimated spatial map $\mathbf{U}_1$. The two variants are therefore successive
relaxations, and Eq.~\eqref{eq:factorised} explains why the spatial variant is
the stronger one. For $k>1$ the same derivation yields a sum of $k$ gated terms,
\begin{equation}
L^{c}_{k}=\mathrm{ReLU}\Big(\textstyle\sum_{i=1}^{k}\sigma_i\,
\mathbf{U}_i\odot\sum_{j=1}^{C}v_i[j]\,A_{j}\Big).
\label{eq:general}
\end{equation}

This analysis yields a compact taxonomy along two independent axes, namely how
channels are combined and which spatial profile is imposed.
Table~\ref{tab:design_space} places Grad-CAM and the two variants within this
design space, and Sec.~\ref{sec:ablation} evaluates each of its cells
experimentally.

\begin{table}[htbp]
\centering
\caption{Design space of gradient-based channel weighting. Rows give the spatial
profile imposed on the weights and columns the way channels are combined.
Grad-CAM occupies the most constrained cell, and the proposed spatial variant
relaxes both constraints. PCG-CAM~\citep{huang2025pcg} also lies
in the uniform row: without its centring step, its channel weights reduce
exactly to $\mathbf{v}_1$.}
\label{tab:design_space}
\small
\begin{tabular}{lcc}
\toprule
 & \textbf{Channel weights: mean} & \textbf{Channel weights: $\mathbf{v}_1$}\\
\midrule
Spatial profile: uniform & Grad-CAM & channel-only (\textsc{svd1})\\
Spatial profile: $\mathbf{u}_1$ & -- & spatial (ours)\\
\bottomrule
\end{tabular}
\end{table}

Eq.~\eqref{eq:svdk} also makes explicit the relation to
HiResCAM~\citep{draelos2020use}, which weights each channel by its raw gradient
map, so that $\mathbf{M}_j$ is column $j$ of $\mathbf{G}$ itself. HiResCAM is
therefore the untruncated limit of Eq.~\eqref{eq:svdk}, reached when $k$ equals
the rank of $\mathbf{G}$, while Grad-CAM corresponds to replacing $\mathbf{G}$
by its projection onto the uniform vector. The rank $k$ thus controls how much of
the spatial structure of the gradient is retained: Grad-CAM keeps none of it,
HiResCAM keeps all of it including its low-energy components, and the proposed
weighting keeps only the dominant modes. At the same layer, this intermediate
choice localises better than either extreme, with a single-layer lift of $2.19$
against $2.02$ for HiResCAM and $1.65$ for Grad-CAM on BRISC
(Table~\ref{tab:loc_brisc}), and $1.72$ against $1.63$ and $1.53$ on VOC
(Table~\ref{tab:loc_voc}).

\paragraph{Sign ambiguity}
\label{sec:signs}
A further property distinguishes the two variants, namely their behaviour under
the sign ambiguity of singular vectors.
Singular vectors are determined only up to a joint sign flip: replacing
$(\mathbf{u}_1,\mathbf{v}_1)$ by $(-\mathbf{u}_1,-\mathbf{v}_1)$ leaves
Eq.~\eqref{eq:svd} unchanged. The rank-$k$ reconstruction
$\widehat{\mathbf{G}}_{k}$, and hence Eq.~\eqref{eq:svdk}, is therefore
invariant to this ambiguity. A map built from $\mathbf{v}_1$ alone, as in
Eq.~\eqref{eq:svd1}, is not: flipping the sign of $\mathbf{v}_1$ negates the map,
and the subsequent ReLU then retains the complement of the intended region.

This is not a theoretical curiosity. Measuring the returned sign against the
direction of the mean gradient, we find it reversed on $100\%$ of BRISC images
and $90.7\%$ of VOC images, since the convention is fixed by the decomposition
routine rather than by the data. Without an explicit correction, the
channel-only variant therefore collapses (Sec.~\ref{sec:ablation}). Restoring it
requires an external reference direction, for which we use the Grad-CAM weights,
whereas the spatial variant needs no such convention. Beyond its finer spatial
resolution, the spatial variant is thus the better-posed of the two
formulations.

\subsection{Resolution-Aware Multi-Scale Fusion}
\label{sec:fusion}

The weighting of Sec.~\ref{sec:svdk} produces one map per layer. This subsection
addresses how such maps should be combined across layers: we first explain how
the set of layers is selected, then derive a criterion that indicates which
resolution levels are needed, and finally specify the rule that fuses the
selected maps.

Most CAM methods explain a single layer, almost always the last convolutional
one, which offers the strongest semantics but the coarsest spatial grid. Methods
that do combine several layers, such as LayerCAM~\citep{jiang2021layercam},
leave the choice of layers unspecified. We argue that this choice determines the
attainable precision of the explanation and should therefore be made explicit.
Depth alone is an unreliable guide for it. In standard CNN backbones the spatial
resolution is reduced only at a few stride-2 transitions, so several
consecutive layers usually share the same resolution. A set formed by taking the
last $n$ layers may therefore contain a single resolution level, in which case
fusion averages redundant maps and adds no spatial detail. We instead select
layers by resolution: the set $\mathcal{S}$ contains at most one layer per
resolution level, taken as the deepest layer of that level so that its semantic
content is preserved.

Selecting by resolution leaves open which resolution levels should be included,
and we argue that the answer depends on the size of the target. A target
covering a fraction $\rho$ of the image occupies approximately
\begin{equation}
n_l(\rho)=\rho\,N_l=\rho\,H_lW_l
\label{eq:cells}
\end{equation}
cells at layer $l$. A map computed at layer $l$ has a single value per cell
before it is upsampled, so it cannot represent detail finer than one cell. The
quantity $n_l$ therefore limits the spatial precision that layer $l$ can
provide. When $n_l$ is close to one or smaller, the whole target falls within a
single cell, and no choice of weights can localise it within that cell.
Eq.~\eqref{eq:cells} thus yields a practical rule: a higher-resolution level
should be included whenever the final layer alone supplies too few cells for the
targets of interest, and the benefit of doing so should shrink as $\rho$ grows.

The threshold at which a second resolution level becomes necessary can be
anticipated from the geometry of the upsampled map. After bilinear upsampling,
a map computed at layer $l$ attains its maximum at the centre of one of its
cells, so the pointing game at that layer succeeds only if the target contains
the centre of the cell that carries the peak. A target whose extent is below
one cell in some direction may lie entirely between two cell centres, in which
case no choice of weights can place the peak inside it; the outcome is then
decided by the alignment of the target with the grid rather than by the
explanation. Robustness to alignment requires the target to span at least two
cells in each direction, since any axis-aligned interval of length two cells
contains a cell centre regardless of its offset, and the same requirement
carries over the stride-two transition that produces the deepest level, where
each cell integrates a $2\times2$ block of the level above. The condition is
therefore $n_l\gtrsim 2\times2=4$ cells, below which the deepest layer alone is
expected to lose the pointing game on a substantial fraction of images
irrespective of the weighting scheme. This is a geometric argument rather than
a theorem, since real targets are not axis-aligned rectangles;
Sec.~\ref{sec:size_law} tests the prediction on two datasets whose targets
differ by more than an order of magnitude in relative size and locates the
measured change of sign between $3.3$ and $4.5$ cells.

Once the layer set has been chosen, the maps of its members must be combined
into a single explanation. To this end, each per-layer map is normalised to
$[0,1]$, upsampled to the input resolution and combined as
\begin{equation}
L^{c}_{\text{fused}}
=\mathcal{N}\Big(\textstyle\sum_{l\in\mathcal{S}}\lambda_l\,
\mathrm{UP}\big(L^{c}_{k}(l)\big)\Big),
\qquad \textstyle\sum_{l}\lambda_l=1,
\label{eq:fusion}
\end{equation}
where $\mathrm{UP}(\cdot)$ is bilinear upsampling and $\mathcal{N}(\cdot)$ is
min--max normalisation to $[0,1]$. Rather than positing a single weighting
scheme, we compare three of them, namely uniform weights, depth-weighted weights
($\lambda_l\!\propto\!D_l$) and inverse-depth weights
($\lambda_l\!\propto\!|\mathcal{S}|\!-\!D_l\!+\!1$), where $D_l$ is the rank of
layer $l$ in $\mathcal{S}$ ordered from shallow to deep. One of them is selected
on a held-out split (Sec.~\ref{sec:protocol}). Consistently with
Eq.~\eqref{eq:cells}, the selected scheme places more weight on the
high-resolution layer for the dataset with smaller targets.

\subsection{Computational Complexity}
\label{sec:complexity}

The method needs one forward and one backward pass, as Grad-CAM does. The only
additional cost is a thin SVD of $\mathbf{G}\in\mathbb{R}^{N\times C}$, i.e.\
$\mathcal{O}\!\left(NC\min(N,C)\right)$, plus $\mathcal{O}(NC)$ for the
element-wise weighting, which is fully vectorised. Both terms remain modest
because $N$ counts feature-map cells rather than pixels, and is typically one to
three orders of magnitude smaller than the input resolution. For a fusion set
$\mathcal{S}$ the costs simply add over its members. We use the exact thin SVD
throughout for reproducibility; a truncated solver returning only the leading
$k$ components would reduce the cost further, since $k\ll\min(N,C)$.

Crucially, the cost is independent of the number of network evaluations. This
contrasts with perturbation-based methods such as Score-CAM, which require one
additional forward pass per channel, i.e.\ $\mathcal{O}(C)$ evaluations.
Measured runtimes are reported in Sec.~\ref{sec:runtime}.

\FloatBarrier
\section{Experiments}
\label{sec:experiments}

We evaluate on two datasets chosen for their contrast in target size, using a
protocol designed so that every claim is falsifiable: the configuration is fixed
on a held-out split before any result is reported, every comparison is paired on
identical images, and effect sizes accompany all significance tests.

\subsection{Experimental Protocol and Reproducibility}
\label{sec:protocol}

This subsection describes the datasets, models and preprocessing, followed by
the protocol used to select the configuration and to test differences between
methods.

We use two datasets that differ markedly in target size. PASCAL VOC
2012~\citep{everingham2010pascal} provides bounding-box annotations for 20 object
categories. For configuration selection we use the first 600 images of its
\texttt{train} list, and for reporting the first 1500 images of its \texttt{val}
list; images in which the predicted class has no annotated box are discarded,
leaving 596 and 1401 images respectively. BRISC 2025~\citep{brisc2025} provides brain-MRI
slices labelled as glioma, meningioma, pituitary tumour or no tumour, together
with pixel-level tumour masks. We use the classification split for training and
the segmentation split for evaluation; the two share the same slices, so masks
are available for every evaluated image. Configuration selection uses 744
tumour slices from the training part of the segmentation split (about 250 per
class), and reporting uses all 859 tumour slices of its test part. Only the three tumour classes enter the
localisation evaluation, since the no-tumour class has no target region.

For VOC we use EfficientNet-B0 and for BRISC EfficientNet-B1~\citep{tan2019efficientnet}, both initialised
from ImageNet weights. The BRISC classifier is fine-tuned in two stages (frozen
backbone, then fine-tuning of its last 30 layers) and reaches $97.90\%$ accuracy on the held-out
test split of 1000 slices, with macro-averaged precision, recall and $F_1$ all
equal to $0.98$; per-class $F_1$ is $0.97$ for glioma, $0.97$ for meningioma,
$1.00$ for the no-tumour class and $0.99$ for pituitary tumours. Explanation
quality is only meaningful for a competent classifier, so we report this
explicitly. All methods are implemented in TensorFlow/Keras, and run times are measured on a single NVIDIA T4 GPU.

Each model is explained under the same preprocessing that was used to train it,
namely brain extraction followed by resizing for BRISC and plain resizing for
VOC. For BRISC the same crop box is applied to the image and to its mask, so that
annotations remain aligned, and we verified that this crop removes no tumour
pixels on any evaluated image. Preprocessing is identical for all compared
methods within a dataset and therefore cannot favour any of them. Target-area
fractions are reported relative to the model input, since that is the quantity
entering Eq.~\eqref{eq:cells}.

The candidate layers for fusion are the deepest convolutional layer at each
distinct spatial resolution. Layers whose output is spatially degenerate, such
as the $1\times1$ tensors inside squeeze-and-excitation blocks, are excluded,
because maps derived from them are constant after upsampling and carry no
localisation information. All single-layer baselines are computed on the deepest
convolutional layer, so that every method is compared under identical
conditions. One consequence is that LayerCAM, which is designed to exploit
shallower layers, is evaluated at a depth that is not its intended operating
point; we keep it on the common layer for fairness of comparison rather than
tuning each baseline to its own most favourable layer.
Among decomposition-based methods we include Eigen-CAM, SVD-Grad-CAM and
PCG-CAM~\citep{huang2025pcg}. Eigen-CAM projects the activations onto their
first principal component and, as in its original formulation, applies no sign
correction. SVD-Grad-CAM decomposes the gradient map
of each channel separately and keeps its leading component. PCG-CAM centres each channel's gradient over space, takes the
leading eigenvector of the spatial covariance computed jointly across channels,
projects the uncentred gradients onto it to obtain channel weights, and applies an
absolute value to the resulting map. Both reduce the decomposition to one scalar
weight per channel, the design choice isolated by the controlled comparison of
Sec.~\ref{sec:ablation}.

The configuration of the proposed method, consisting of the resolution levels
to fuse, the fusion weighting and the rank $k$, is selected on the training split
of each dataset using a composite of the three localisation metrics, and is then
held fixed. All reported results are obtained on the evaluation split with that
fixed configuration, so no hyperparameter is tuned on the data used for
reporting.

To compare methods, all tests are paired on identical images. We use the
Wilcoxon signed-rank test and report, alongside the $p$-value, the matched-pairs
rank-biserial correlation as an effect size and the fraction of images on which
the proposed method wins. Because the metric distributions are strongly skewed,
central tendency is reported as the median with the interquartile range for
EBPG and lift, and as the mean for the pointing game and IoU, which are
bounded. All localisation conclusions drawn from paired tests remain valid
after Holm--Bonferroni correction over the full set of paired comparisons
reported in Tables~\ref{tab:paired_brisc} and~\ref{tab:paired_voc}.

Evaluation code, configuration files and the scripts reproducing the tables and
figures are publicly available at
\url{https://github.com/newshaShah/SVD_Fusion_CAM}. Both datasets are publicly
available at their original sources: PASCAL VOC 2012 from the challenge website,
and BRISC 2025 from its published repository~\citep{brisc2025}.

\subsection{Evaluation Metrics}
\label{sec:metrics}

We assess each explanation from three angles: how well it localises the
annotated target, how faithfully it reflects the evidence used by the model, and
how strongly it depends on the learned parameters.

For localisation, the energy-based pointing game
(EBPG)~\citep{wang2020scorecam} is the fraction of the normalised saliency mass
that falls inside the annotated region. Because its value depends on how large
that region is, we also report the lift, defined as the ratio of EBPG to the
target's area fraction, so that a value of one corresponds to a uniform map and
values are comparable across datasets with very different target sizes. The
pointing game (PG)~\citep{zhang2018top} records whether the maximum of the map
lies inside the annotation, and IoU measures the overlap between the annotation
and the map thresholded at half its maximum; \ref{app:iou_thr} reports IoU
at thresholds of $0.3$ and $0.7$ as well.

For faithfulness, deletion and insertion curves~\citep{petsiuk2018rise} remove
or restore pixels in the order given by the map and track the class score, and
we report the area under each curve. We do not report average drop or average
increase, following the analysis of \citet{zhang2024opticam}, who identify a
flaw in both.

For model dependence, cascading weight randomisation~\citep{adebayo2018sanity}
progressively replaces the trained weights, starting from the classifier head,
and measures how much the explanation changes. A method whose output is
unaffected by this randomisation is not explaining the model.

\subsection{Configuration Selection and Ablation}
\label{sec:ablation}

This subsection first reports how the configuration of the proposed method was
selected, and then isolates the contribution of each of its design choices.
Following the resolution-based selection of Sec.~\ref{sec:fusion}, the candidate
levels are the distinct spatial resolutions available in the backbone, each
represented by its deepest layer. For each dataset we evaluated a set of
combinations of these levels, the three fusion weightings of Sec.~\ref{sec:fusion}, and
$k \in \{1,2,3\}$, scoring each configuration on the training split by the
normalised mean of lift, pointing game and IoU. The selected configurations are
reported in Table~\ref{tab:config} and were then held fixed for all subsequent
results.

\begin{table}[htbp]
\centering
\caption{Configurations selected on the training split of each dataset and used
unchanged for all reported results. Both consist of one mid-resolution level and
the deepest level, but the weight assigned to the higher-resolution level is
larger for the dataset with smaller targets.}
\label{tab:config}
\small
\begin{tabular}{lccc}
\toprule
\textbf{Dataset} & \textbf{Resolution levels} & \textbf{Rank $k$} & \textbf{Fusion weighting}\\
\midrule
BRISC 2025 & $30\times30$, $8\times8$ & 3 & uniform (0.50 / 0.50)\\
PASCAL VOC 2012 & $28\times28$, $7\times7$ & 1 & depth (0.33 / 0.67)\\
\bottomrule
\end{tabular}
\end{table}

Two observations follow. The selected sets have the same structure on both
datasets, namely one mid-resolution level combined with the deepest one, although
they were chosen independently and on different backbones. And the selected
weighting differs in the direction predicted by Eq.~\eqref{eq:cells}: the dataset
whose targets are smaller assigns half the weight to the high-resolution level,
whereas the dataset with large targets assigns only a third. VOC still benefits
from fusion despite its large mean object size because a substantial fraction of
its objects are small (Sec.~\ref{sec:size_law}); the smaller high-resolution
weight reflects that the deepest layer already suffices for the majority.

Having fixed the configuration, we next examine the two design axes of
Sec.~\ref{sec:svdk} in isolation, namely how channels are combined and which
spatial profile is imposed. Table~\ref{tab:taxonomy}
evaluates each cell of that design space on the same backbone, the same layer and
the same images, so that no other factor varies.

\begin{table}[htbp]
\centering
\caption{The two design axes evaluated in isolation. All rows use the deepest
convolutional layer only, with no fusion, on the full evaluation splits
($n=859$ BRISC and $n=1401$ VOC images). Row~2 uses the principal channel mode
as returned by the decomposition and row~3 additionally aligns its sign with
the mean gradient direction; images on which the uncorrected map is
identically zero (1 and 85 respectively) are excluded from row~2.}
\label{tab:taxonomy}
\resizebox{\linewidth}{!}{
\begin{tabular}{llcccccc}
\toprule
& & \multicolumn{3}{c}{\textbf{BRISC 2025}} & \multicolumn{3}{c}{\textbf{PASCAL VOC 2012}}\\
\cmidrule(lr){3-5}\cmidrule(lr){6-8}
\textbf{Channel weights} & \textbf{Spatial profile} & Lift & PG (\%) & IoU (\%) & Lift & PG (\%) & IoU (\%)\\
\midrule
mean gradient (Grad-CAM) & uniform & 1.65 & 19.91 & 11.77 & 1.53 & 89.01 & \textbf{37.75}\\
$\mathbf{v}_1$, raw sign & uniform & 0.02 & 0.93 & 0.87 & 0.38 & 31.23 & 5.25\\
$\mathbf{v}_1$, sign-corrected & uniform & 1.87 & 22.24 & 13.00 & 1.60 & 90.51 & 37.18\\
$\mathbf{v}_1$ & $\mathbf{u}_1$ \textbf{(ours)} & \textbf{2.19} & \textbf{25.84} & \textbf{14.92} & \textbf{1.72} & \textbf{91.15} & 29.01\\
\bottomrule
\end{tabular}}
\end{table}

Three conclusions follow from Table~\ref{tab:taxonomy}.

First, the sign ambiguity of the singular vectors is a practical obstacle, not a
theoretical one. The sign returned by the decomposition was opposed to the mean
gradient direction on $100\%$ of BRISC images and $90.7\%$ of VOC images, and the
uncorrected channel-only variant consequently retains the complement of the
intended region and performs at or below chance. Correcting it requires an
external reference direction. The proposed formulation requires no such
convention because the rank-$k$ reconstruction is invariant to the ambiguity.

Second, both axes contribute, and the spatial axis contributes more. Replacing
the mean gradient by the principal channel mode raises lift by $0.22$ on BRISC
and $0.07$ on VOC; adding the spatial gate raises it by a further $0.32$ and
$0.12$. The spatial axis accounts for $59\%$ and $63\%$ of the total improvement
over Grad-CAM on the two datasets respectively, which is the empirical
counterpart of Eq.~\eqref{eq:gradcam_error}: the gain comes from estimating the
spatial factor rather than fixing it.

Third, the improvement is larger where targets are smaller: $+0.54$ lift on BRISC
against $+0.19$ on VOC, consistent with Sec.~\ref{sec:size_law}. On VOC the
single-layer pointing game improves only modestly along the two axes
($+2.1$ points); the larger gain reported in Table~\ref{tab:loc_voc} comes from
fusion rather than from the weighting.

Beyond the weighting itself, a second concern is whether the improvement stems from the proposed weighting or
merely from aggregating more than one layer. Table~\ref{tab:2x2} applies the
identical fusion rule to both weighting schemes.

\begin{table}[htbp]
\centering
\caption{Fusion applied identically to the scalar weighting of SVD-Grad-CAM and
to the proposed spatial weighting, on the full evaluation splits of BRISC 2025
($n=859$) and PASCAL VOC 2012 ($n=1401$). The comparison of interest is single
versus fused within each weighting, not across rows. SVD-Grad-CAM is used as
the scalar control because it shares the spectral decomposition with our
method and differs only in reducing each channel to a scalar, so the comparison
isolates the spatial weight map. The best value in each column is shown in
bold. Fusion is not beneficial in isolation: with the scalar weighting it
lowers every metric on BRISC and lowers lift and IoU on VOC, whereas with the
spatial weighting it raises the pointing game on both datasets ($+17.5$ and
$+3.6$ points) and, on BRISC, lift and IoU as well. The higher IoU of the
scalar method on VOC reflects the same large-object effect as in
Table~\ref{tab:loc_voc}: at a fixed threshold a diffuse map overlaps a large
box more fully than a focused one.}
\label{tab:2x2}
\resizebox{\linewidth}{!}{
\begin{tabular}{llcccccc}
\toprule
& & \multicolumn{3}{c}{\textbf{BRISC 2025}} & \multicolumn{3}{c}{\textbf{PASCAL VOC 2012}}\\
\cmidrule(lr){3-5}\cmidrule(lr){6-8}
\textbf{Weighting} & \textbf{Layers} & Lift & PG (\%) & IoU (\%) & Lift & PG (\%) & IoU (\%)\\
\midrule
scalar (SVD-Grad-CAM) & single & 1.63 & 20.49 & 11.71 & 1.54 & 90.01 & \textbf{37.45}\\
scalar (SVD-Grad-CAM) & fused & 1.47 & 18.74 & 10.04 & 1.46 & 90.51 & 35.05\\
\textbf{spatial (ours)} & single & 2.19 & 25.84 & 14.92 & \textbf{1.72} & 91.15 & 29.01\\
\textbf{spatial (ours)} & fused & \textbf{2.31} & \textbf{43.31} & \textbf{15.42} & 1.71 & \textbf{94.72} & 26.10\\
\bottomrule
\end{tabular}}
\end{table}

Fusion changes the pointing game on BRISC by $-1.8$ points when combined with
scalar weighting and by $+17.5$ points when combined with spatial weighting. It is
therefore not an additive component that could be applied to any method, but one
that pays off only in combination with a spatially resolved weight map: a coarse
map from a deeper layer and a coarse map from a shallower layer average into a
blurrier result, whereas two spatially resolved maps reinforce one another at the
location of the peak. This addresses the question of whether the gain originates
in the weighting, in the aggregation, or in the experimental setup: the setup is
identical across all four rows, aggregation alone is harmful, and only the
combination improves the pointing game, while lift and IoU change by at most
a few tenths of a point. The benefit of fusion is therefore mainly a gain in
peak placement, which is the quantity that the cell-coverage argument of
Eq.~\eqref{eq:cells} concerns (Sec.~\ref{sec:size_law}).

Finally, Table~\ref{tab:ksens} reports the sensitivity of the selected configuration to
$k$. The method is robust to this choice: lift varies by $2.2\%$ relative to $k=1$ on VOC
and $7.5\%$ on BRISC across $k \in \{1,2,3\}$, which is small against the
margins between methods in Tables~\ref{tab:loc_brisc} and~\ref{tab:loc_voc}.
The selected values, $k=1$ on VOC and $k=3$ on BRISC, reflect the slightly
higher gradient complexity of the medical images; either value gives the
same ranking of methods. Extending the range to $k=5$ on the evaluation
split changes lift by at most $3\%$ ($2.31$ against $2.31$ on BRISC, $1.66$
against $1.71$ on VOC) and the pointing game by less than one point
(\ref{app:energy}). The retained energy $\eta_k$ reported there explains why
no energy threshold can replace this choice: the spectrum decays slowly, so
any threshold above $0.7$ selects the largest rank considered, and the method
is simply insensitive to $k$ within this range.

\begin{table}[htbp]
\centering
\caption{Sensitivity to the rank $k$, measured on the training split for the
selected layer set of each dataset.}
\label{tab:ksens}
\begin{tabular}{lcccccc}
\toprule
& \multicolumn{3}{c}{\textbf{BRISC 2025} ($30{+}8$)} & \multicolumn{3}{c}{\textbf{PASCAL VOC 2012} ($28{+}7$)}\\
\cmidrule(lr){2-4}\cmidrule(lr){5-7}
\textbf{$k$} & Lift & PG (\%) & IoU (\%) & Lift & PG (\%) & IoU (\%)\\
\midrule
1 & 2.239 & 45.57 & 15.84 & \textbf{1.711} & 93.46 & 25.98\\
2 & 2.132 & 45.97 & 16.19 & 1.676 & 92.79 & \textbf{26.65}\\
3 & \textbf{2.301} & \textbf{47.98} & \textbf{17.03} & 1.673 & \textbf{94.13} & 26.51\\
\bottomrule
\end{tabular}
\end{table}

\subsection{Comparison with State-of-the-Art Methods}
\label{sec:sota}

Tables~\ref{tab:loc_brisc} and~\ref{tab:loc_voc} report localisation performance
on both datasets, and Tables~\ref{tab:paired_brisc} and~\ref{tab:paired_voc} the
corresponding paired comparisons.

\begin{table}[htbp]
\centering
\caption{Localisation performance on the BRISC 2025 test split ($n=859$ tumour
images). Lift is EBPG divided by the fraction of the image the tumour occupies,
so a value of $1$ corresponds to chance. EBPG and lift are reported as the
median with the interquartile range because their distributions are strongly
skewed; PG and IoU are means over images.}
\label{tab:loc_brisc}
\resizebox{\linewidth}{!}{
\begin{tabular}{lcccc}
\toprule
\textbf{Method} & \textbf{EBPG (\%)} & \textbf{Lift} & \textbf{PG (\%)} $\uparrow$ & \textbf{IoU (\%)} $\uparrow$\\
 & median [IQR] & median [IQR] & mean & mean \\
\midrule
\textbf{SVD-Fusion CAM (multi-scale)} & 4.8 [2.0--13.8] & \textbf{2.31 [1.36--3.80]} & \textbf{43.31} & \textbf{15.42}\\
\textbf{SVD-Fusion CAM (single layer)} & 4.5 [1.7--13.1] & 2.19 [1.24--3.80] & 25.84 & 14.92\\
HiResCAM~\citep{draelos2020use} & 4.0 [1.5--11.9] & 2.02 [1.09--3.38] & 25.38 & 14.36\\
Grad-CAM++~\citep{chattopadhay2018gradcam} & 3.9 [0.7--11.4] & 1.76 [0.60--2.90] & 23.63 & 12.80\\
LayerCAM~\citep{jiang2021layercam} & 3.8 [1.8--8.6] & 1.67 [1.37--2.15] & 30.73 & 14.66\\
Grad-CAM~\citep{selvaraju2017grad} & 3.4 [1.0--9.8] & 1.65 [0.76--2.83] & 19.91 & 11.77\\
PCG-CAM~\citep{huang2025pcg} & 3.3 [1.5--8.6] & 1.64 [1.03--2.39] & 24.56 & 12.77\\
SVD-Grad-CAM~\citep{thota2024svd} & 3.4 [1.0--9.5] & 1.63 [0.75--2.82] & 20.49 & 11.71\\
XGrad-CAM~\citep{fu2020axiom} & 1.3 [0.1--5.0] & 0.65 [0.09--2.00] & 7.22 & 4.19\\
Eigen-CAM~\citep{muhammad2020eigen} & 1.0 [0.0--6.5] & 0.56 [0.00--2.68] & 12.22 & 6.24\\
\bottomrule
\end{tabular}}
\end{table}

On the medical dataset the proposed method leads on all three metrics. The
strongest baseline overall is HiResCAM, over which the margin is $14.4\%$ in
lift (2.31 against 2.02) and $71\%$ in the pointing game (43.31 against
25.38); the strongest baseline on the pointing game alone is LayerCAM
(30.73), over which the margin is $41\%$, although Sec.~\ref{sec:sanity} shows
that LayerCAM's map is largely insensitive to the classifier weights. The
paired tests confirm that these margins are not driven by a minority of
images: on lift the proposed method wins on $73$--$86\%$ of images depending
on the baseline, and rank-biserial effect sizes across lift and the pointing
game lie between $0.44$ and $0.88$. The one comparison that is not significant
is IoU against LayerCAM ($p = 0.87$), so we claim parity rather than
superiority there. PCG-CAM, the most recent decomposition-based baseline,
improves on Grad-CAM in the pointing game and IoU but not in lift ($1.64$
against $1.65$), and the proposed method exceeds it in lift on $78.9\%$ of
images ($r=0.76$).

XGrad-CAM performs below chance on this dataset ($\text{lift}=0.65$). Its
derivation normalises the gradient weights by the summed channel activation and
assumes non-negative activations, which holds for ReLU networks but not for the
Swish activations of the backbone used here, where the pre-activation tensor we
explain can take negative values and the normalising term becomes ill-conditioned.
We report the measurement rather than omitting it, but do not treat it as
evidence about the method in the setting for which it was derived.

\begin{table}[p]
\centering
\caption{Paired comparison on BRISC 2025 ($n=859$ tumour images) between
SVD-Fusion CAM (multi-scale) and every other method, using the Wilcoxon
signed-rank test on identical images. Lift, the pointing game and IoU are the localisation metrics of
Sec.~\ref{sec:metrics}. Win rate is the percentage of images on which the
proposed method scores strictly higher, and the effect size $r$ is the
matched-pairs rank-biserial correlation, which ranges from $-1$ to $+1$ and is
positive when the proposed method is better. Large effects in favour of the
proposed method ($r\ge 0.5$) are shown in bold.}
\label{tab:paired_brisc}
\resizebox{\linewidth}{!}{
\begin{tabular}{llccc}
\toprule
\textbf{Metric} & \textbf{Compared with} & \textbf{Win rate (\%)} & \textbf{Effect size $r$} & \textbf{$p$}\\
\midrule
\multirow{9}{*}{Lift}
 & SVD-Fusion CAM (single layer) & 59.5 & $+0.291^{*}$ & 1.6e-13\\
 & HiResCAM & 75.4 & $\mathbf{+0.622}^{*}$ & 4.1e-56\\
 & Grad-CAM++ & 75.2 & $\mathbf{+0.656}^{*}$ & 3.3e-62\\
 & LayerCAM & 73.1 & $\mathbf{+0.700}^{*}$ & 1.3e-70\\
 & Grad-CAM & 83.9 & $\mathbf{+0.837}^{*}$ & 2.7e-100\\
 & PCG-CAM & 78.9 & $\mathbf{+0.762}^{*}$ & 2.7e-83\\
 & SVD-Grad-CAM & 85.6 & $\mathbf{+0.857}^{*}$ & 6.7e-105\\
 & XGrad-CAM & 77.9 & $\mathbf{+0.676}^{*}$ & 4.1e-66\\
 & Eigen-CAM & 74.6 & $\mathbf{+0.603}^{*}$ & 6.5e-53\\
\midrule
\multirow{9}{*}{Pointing game}
 & SVD-Fusion CAM (single layer) & 22.6 & $\mathbf{+0.630}^{*}$ & 2.4e-22\\
 & HiResCAM & 22.2 & $\mathbf{+0.675}^{*}$ & 2.0e-24\\
 & Grad-CAM++ & 24.8 & $\mathbf{+0.658}^{*}$ & 5.5e-26\\
 & LayerCAM & 20.5 & $+0.443^{*}$ & 4.7e-12\\
 & Grad-CAM & 26.5 & $\mathbf{+0.788}^{*}$ & 2.5e-36\\
 & PCG-CAM & 24.4 & $\mathbf{+0.622}^{*}$ & 1.5e-23\\
 & SVD-Grad-CAM & 26.2 & $\mathbf{+0.772}^{*}$ & 9.3e-35\\
 & XGrad-CAM & 38.6 & $\mathbf{+0.876}^{*}$ & 5.4e-61\\
 & Eigen-CAM & 34.8 & $\mathbf{+0.807}^{*}$ & 9.2e-49\\
\midrule
\multirow{9}{*}{IoU}
 & SVD-Fusion CAM (single layer) & 52.4 & $+0.155^{*}$ & 2.2e-04\\
 & HiResCAM & 51.9 & $+0.174^{*}$ & 3.5e-05\\
 & Grad-CAM++ & 56.1 & $+0.243^{*}$ & 5.2e-09\\
 & LayerCAM & 44.4 & $+0.007$ & 8.7e-01\\
 & Grad-CAM & 59.6 & $+0.401^{*}$ & 1.3e-21\\
 & PCG-CAM & 53.3 & $+0.237^{*}$ & 8.4e-09\\
 & SVD-Grad-CAM & 60.0 & $+0.407^{*}$ & 3.5e-22\\
 & XGrad-CAM & 71.4 & $\mathbf{+0.688}^{*}$ & 3.8e-61\\
 & Eigen-CAM & 66.5 & $\mathbf{+0.586}^{*}$ & 7.4e-45\\
\bottomrule
\end{tabular}}
\begin{minipage}{\linewidth}\vspace{2pt}\footnotesize
$^{*}$ significant at $p<0.05$. For the binary pointing game most images are
ties, so the win rate should be read against the complementary loss rate rather
than against $50\%$.
\end{minipage}
\end{table}

\begin{table}[htbp]
\centering
\caption{Localisation performance on the PASCAL VOC 2012 validation split
($n=1401$). Ground truth comprises the boxes of the predicted class. Objects
occupy $44.4\%$ of the image on average; for an object of this size, even a
perfectly concentrated map cannot exceed a lift of $1/0.444\approx2.25$. EBPG
and lift are medians with interquartile range; PG and IoU are means.}
\label{tab:loc_voc}
\resizebox{\linewidth}{!}{
\begin{tabular}{lcccc}
\toprule
\textbf{Method} & \textbf{EBPG (\%)} & \textbf{Lift} & \textbf{PG (\%)} $\uparrow$ & \textbf{IoU (\%)} $\uparrow$\\
 & median [IQR] & median [IQR] & mean & mean \\
\midrule
\textbf{SVD-Fusion CAM (multi-scale)} & 78.3 [58.3--89.7] & 1.71 [1.27--2.75] & \textbf{94.72} & 26.10\\
\textbf{SVD-Fusion CAM (single layer)} & 79.0 [59.0--90.3] & \textbf{1.72 [1.27--2.77]} & 91.15 & 29.01\\
HiResCAM~\citep{draelos2020use} & 73.0 [54.7--85.4] & 1.63 [1.22--2.53] & 91.36 & 34.81\\
SVD-Grad-CAM~\citep{thota2024svd} & 71.5 [48.9--85.4] & 1.54 [1.18--2.33] & 90.01 & 37.45\\
Grad-CAM~\citep{selvaraju2017grad} & 71.2 [48.2--85.2] & 1.53 [1.16--2.30] & 89.01 & 37.75\\
PCG-CAM~\citep{huang2025pcg} & 64.4 [42.0--80.5] & 1.42 [1.12--2.07] & 86.30 & 31.36\\
Grad-CAM++~\citep{chattopadhay2018gradcam} & 64.5 [39.2--81.3] & 1.40 [1.14--1.92] & 88.22 & \textbf{42.75}\\
LayerCAM~\citep{jiang2021layercam} & 60.9 [38.8--77.9] & 1.37 [1.13--1.85] & 84.80 & 39.77\\
XGrad-CAM~\citep{fu2020axiom} & 54.9 [23.7--79.2] & 1.14 [0.89--1.52] & 61.31 & 14.98\\
Eigen-CAM~\citep{muhammad2020eigen} & 49.4 [13.4--85.6] & 1.02 [0.41--1.73] & 55.54 & 17.12\\
\bottomrule
\end{tabular}}
\end{table}

On natural images the picture differs in three respects, each of which is
predicted by the analysis of Sec.~\ref{sec:fusion}. Both variants of the
proposed method outperform every baseline on EBPG and lift, with win rates
against baselines ranging from $69\%$ to $97\%$; the margin over the strongest
baseline is smaller here ($4.9\%$ in lift over HiResCAM) than on the medical
dataset, as expected for targets that occupy many cells. Second, the
multi-scale variant no longer improves on the single-layer one in energy
concentration: on lift and EBPG the single-layer variant is significantly
better (effect sizes $-0.23$ and $-0.32$), while the pointing game favours
fusion ($+3.6$ points, $p = 3.8\text{e-}9$). This is precisely what
Eq.~\eqref{eq:cells} predicts when targets are large, and we analyse it in
Sec.~\ref{sec:size_law}.

Third, both variants score below most baselines on IoU. This is a property of
the metric under this annotation rather than of the maps: with ground truth
covering $44\%$ of the image on average, a diffuse map that spreads mass
broadly is rewarded by a fixed-threshold overlap measure, whereas a
concentrated map is penalised. The ranking reverses for small objects: below
$10\%$ of the image the proposed method attains the highest IoU of all methods
($33.3\%$ against $31.2\%$ for PCG-CAM, the next best), and below $5\%$ the
margin widens ($28.0\%$ against $23.1\%$). \ref{app:iou_thr} repeats the
comparison at thresholds of $0.3$ and $0.7$: for small objects the proposed
method remains the best at $0.3$ and $0.5$, whereas at $0.7$, where only the
immediate neighbourhood of the peak survives thresholding, the compact fused
map covers less of the box than the broader maps of LayerCAM and HiResCAM.
The paired VOC comparisons are reported in \ref{app:paired_voc}.

\subsection{Per-Class Analysis}
\label{sec:perclass}

The three tumour types differ both in how much of the image they occupy and in
how well any method localises them, and the two do not correspond in the way one
might expect. Table~\ref{tab:perclass} reports both together.

\begin{table}[p]
\centering
\caption{Per-class results on BRISC 2025. The upper block reports the mean
tumour area and the number of cells it occupies at the deepest layer
($8\times8$). Cell coverage bounds the attainable precision but does not by
itself predict performance: pituitary tumours are sub-cell and are hard for every
single-layer method on the pointing game, whereas gliomas and meningiomas occupy a similar number of cells yet
differ by more than a factor of two in lift.}
\label{tab:perclass}
\resizebox{\linewidth}{!}{
\begin{tabular}{lccc}
\toprule
& \textbf{Glioma} & \textbf{Meningioma} & \textbf{Pituitary}\\
\midrule
Images & 254 & 305 & 300\\
Mean tumour area (\%) & 3.86 & 4.66 & 1.41\\
Cells at deepest layer ($8\times8$) & 2.47 & 2.98 & \textbf{0.90}\\
Cells at fused layer ($30\times30$) & 34.7 & 41.9 & 12.7\\
\midrule
\multicolumn{4}{l}{\textit{Lift (median)}}\\
\textbf{SVD-Fusion CAM (multi-scale)} & 1.73 & \textbf{4.20} & \textbf{1.70}\\
\textbf{SVD-Fusion CAM (single layer)} & 1.71 & 4.16 & 1.63\\
HiResCAM~\citep{draelos2020use} & 1.35 & 3.74 & 1.61\\
SVD-Grad-CAM~\citep{thota2024svd} & 1.26 & 2.97 & 0.89\\
Grad-CAM~\citep{selvaraju2017grad} & 1.30 & 3.01 & 0.84\\
PCG-CAM~\citep{huang2025pcg} & 1.36 & 2.29 & 1.46\\
Grad-CAM++~\citep{chattopadhay2018gradcam} & \textbf{1.85} & 3.15 & 0.38\\
LayerCAM~\citep{jiang2021layercam} & 1.50 & 2.21 & 1.54\\
XGrad-CAM~\citep{fu2020axiom} & 0.62 & 0.68 & 0.66\\
Eigen-CAM~\citep{muhammad2020eigen} & 0.69 & 1.03 & 0.25\\
\midrule
\multicolumn{4}{l}{\textit{Pointing game (\%)}}\\
\textbf{SVD-Fusion CAM (multi-scale)} & \textbf{23.6} & \textbf{75.1} & \textbf{27.7}\\
\textbf{SVD-Fusion CAM (single layer)} & 14.6 & 53.4 & 7.3\\
HiResCAM~\citep{draelos2020use} & 12.2 & 55.7 & 5.7\\
SVD-Grad-CAM~\citep{thota2024svd} & 9.8 & 47.9 & 1.7\\
Grad-CAM~\citep{selvaraju2017grad} & 9.4 & 46.9 & 1.3\\
PCG-CAM~\citep{huang2025pcg} & 16.9 & 49.8 & 5.3\\
Grad-CAM++~\citep{chattopadhay2018gradcam} & 15.4 & 53.1 & 0.7\\
LayerCAM~\citep{jiang2021layercam} & 20.1 & 57.7 & 12.3\\
XGrad-CAM~\citep{fu2020axiom} & 4.7 & 12.1 & 4.3\\
Eigen-CAM~\citep{muhammad2020eigen} & 6.3 & 23.3 & 6.0\\
\bottomrule
\end{tabular}}
\end{table}

The cell-coverage criterion accounts for the most striking per-class result,
the difficulty of pituitary tumours. These tumours occupy $1.41\%$ of the image on average, which corresponds to
$0.90$ cells at the deepest layer. By Eq.~\eqref{eq:cells} this is the sub-cell
regime, in which the peak of a map from that layer rarely falls inside the
target, irrespective of how the channel weights are computed. The measurements agree: five of the eight
baselines fall below chance in lift on this class, and Grad-CAM++ places its
maximum inside the tumour in $0.7\%$ of cases. The failure is predicted rather
than incidental.

The criterion does not, however, explain every difference between classes.
Gliomas and meningiomas occupy a comparable number of cells ($2.47$ and $2.98$),
yet the median lift differs by more than a factor of two ($1.73$ against
$4.20$). Cell coverage therefore behaves as an upper bound on what can be
resolved, not as a predictor of what is resolved. The residual difference must
originate in the evidence the classifier uses. Meningiomas are circumscribed
extra-axial masses with sharp margins and fairly uniform enhancement, so the
discriminative evidence largely coincides with the annotated region. Gliomas are
infiltrative, with indistinct margins and surrounding oedema; a classifier may
rely on mass effect and peritumoural signal change, both of which lie outside the
segmentation mask. Under that reading a faithful explanation would be penalised
by a localisation metric, which is the limitation discussed in
Sec.~\ref{sec:related}.

The per-class results also show where multi-scale fusion contributes most, and
this is exactly where the criterion says it should. Adding
the $30\times30$ level raises the coverage of pituitary tumours from $0.90$ to
$12.7$ cells, and the pointing game on that class rises from $7.3\%$ to $27.7\%$,
a factor of $3.8$ over the single-layer variant and more than twice the best
baseline ($12.3\%$ for LayerCAM). The gain is smaller on the two classes that
already have adequate coverage. In terms of lift, the proposed method is the
best on two of the three classes.

The exception is the glioma class, on which Grad-CAM++ attains a slightly higher
median lift ($1.85$ against $1.73$), although its pointing game on this class is
lower ($15.4\%$ against $23.6\%$), indicating a map that is
better centred on average but less often peaked inside the lesion.

\subsection{Target Size and the Benefit of Fusion}
\label{sec:size_law}

Eq.~\eqref{eq:cells} predicts that multi-scale fusion should help in proportion
to how few cells the target occupies at the deepest layer, and the geometric
argument of Sec.~\ref{sec:fusion} places the transition near four cells. We
test both by partitioning VOC into disjoint size bins and treating BRISC as an
independent point.

\begin{table}[htbp]
\centering
\caption{Advantage of the proposed method as a function of target size. VOC rows
are disjoint size bins, so each row is computed on a different set of images; the
BRISC rows are the corresponding bins of the tumour set. The strongest baseline
is HiResCAM in every bin. Gain is the relative improvement in lift over that
baseline, $(\text{lift}_{\text{ours}}-\text{lift}_{\text{baseline}})/
\text{lift}_{\text{baseline}}$. The two BRISC tumours larger than $20\%$ of the image are not shown as a
separate bin. The advantage is larger for small targets than for
large ones.}
\label{tab:size_law}
\resizebox{\linewidth}{!}{
\begin{tabular}{llccccc}
\toprule
\textbf{Dataset} & \textbf{Size bin} & \textbf{Mean area (\%)} & \textbf{$n$} & \textbf{Lift (ours)} & \textbf{Best baseline} & \textbf{Gain (\%)}\\
\midrule
BRISC & 0--5\% & 1.95 & 667 & 2.11 & 1.82 & $+15.8$\\
BRISC & 5--10\% & 6.72 & 155 & 3.76 & 3.48 & $+8.1$\\
BRISC & 10--20\% & 12.48 & 35 & 2.99 & 2.80 & $+7.0$\\
\midrule
VOC & 0--5\% & 2.48 & 103 & 6.22 & 5.57 & $+11.5$\\
VOC & 5--10\% & 7.39 & 83 & 4.95 & 4.10 & $+20.6$\\
VOC & 10--20\% & 15.19 & 144 & 3.45 & 3.07 & $+12.6$\\
VOC & 20--35\% & 27.23 & 254 & 2.44 & 2.27 & $+7.4$\\
VOC & 35--100\% & 63.94 & 817 & 1.34 & 1.28 & $+4.5$\\
\bottomrule
\end{tabular}}
\end{table}

\begin{figure}[htbp]
\centering
\includegraphics[width=\linewidth]{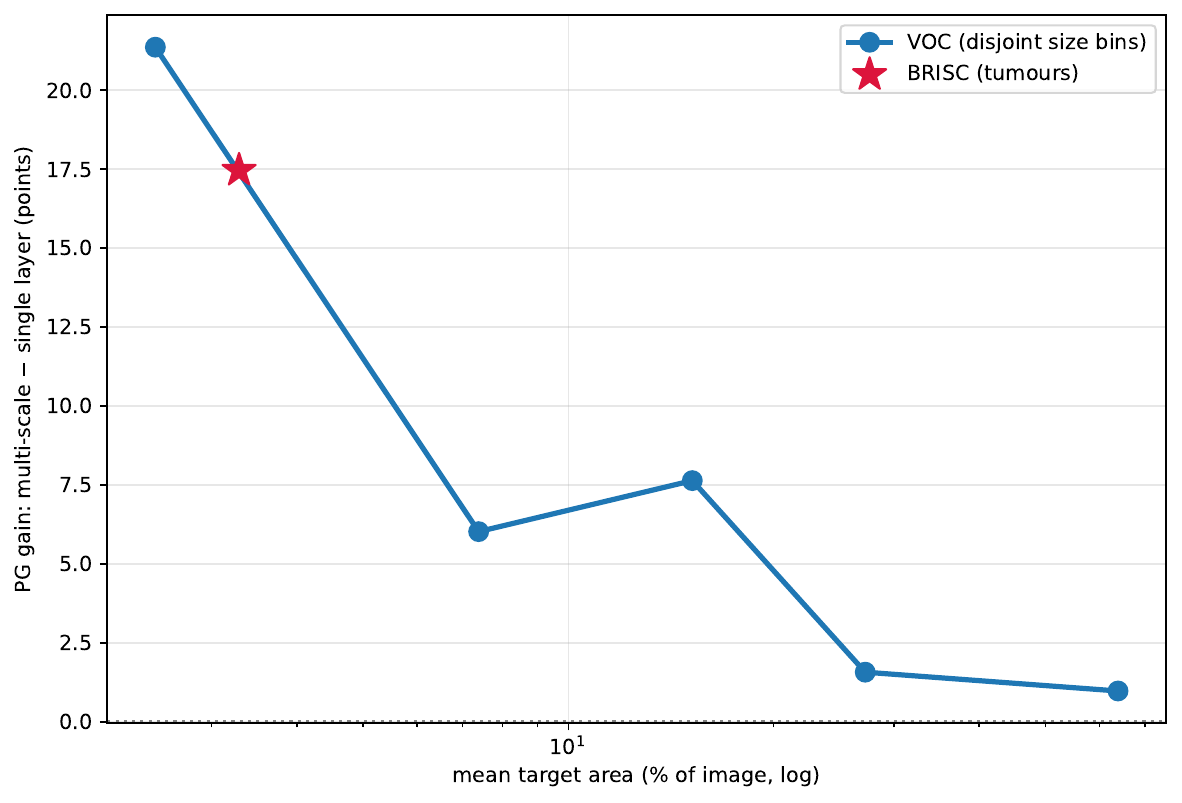}
\caption{Gain in pointing game from multi-scale fusion relative to the
single-layer variant, as a function of target size. VOC points are disjoint size
bins, so each is computed on a different set of images; the BRISC point is the
full tumour set, placed at its measured mean area, and is shown as a single point
because most BRISC tumours fall within the smallest VOC bin. The horizontal axis is
logarithmic; smaller targets are on the left, where the fusion gain is largest.
The quantity does not depend on any baseline, so it is unaffected by the choice
of comparison methods.}
\label{fig:advantage_size}
\end{figure}

Two quantities behave differently and we separate them. The gain contributed by
fusion, shown in Fig.~\ref{fig:advantage_size}, falls from $21.4$ points for
objects below $5\%$ of the image to $1.0$ point for objects above a third of it.
This quantity compares two variants of our own method, multi-scale against
single-layer, and is therefore independent of which baselines are included.

The advantage over the strongest baseline, reported in Table~\ref{tab:size_law},
is also larger for small targets than for large ones ($+8.1\%$ to
$+20.6\%$ below a tenth of the image against $+4.5\%$ above a third). It is
not strictly monotone across bins, because the strongest baseline, HiResCAM,
is itself comparatively strong on the smallest objects; we therefore rely on
the baseline-independent quantity for the claim that follows from
Eq.~\eqref{eq:cells}. \ref{app:per_size} gives the absolute lift of
the methods in each bin, from which the identity of the strongest baseline in
each stratum can be read off.
On VOC, an image-level Spearman rank correlation between target area and the
pointing-game gain of fusion, which does not depend on any binning, is negative
and highly significant
($\rho_s = -0.158$, $p = 3.1\text{e-}9$), confirming the trend without
reference to the bins.

The BRISC measurement was not used to construct this relationship, and provides
an independent check. Interpolating the VOC trend logarithmically to the measured
BRISC mean area of $3.29\%$ predicts a fusion gain of about $17.4$ points,
close to the measured value of $17.5$ points. Given the uncertainty of
estimates obtained from size bins, this agreement should not be over-interpreted,
but it is consistent with a relationship that depends on target size rather than
on the domain, backbone or annotation type.

\begin{figure}[htbp]
\centering
\includegraphics[width=\linewidth]{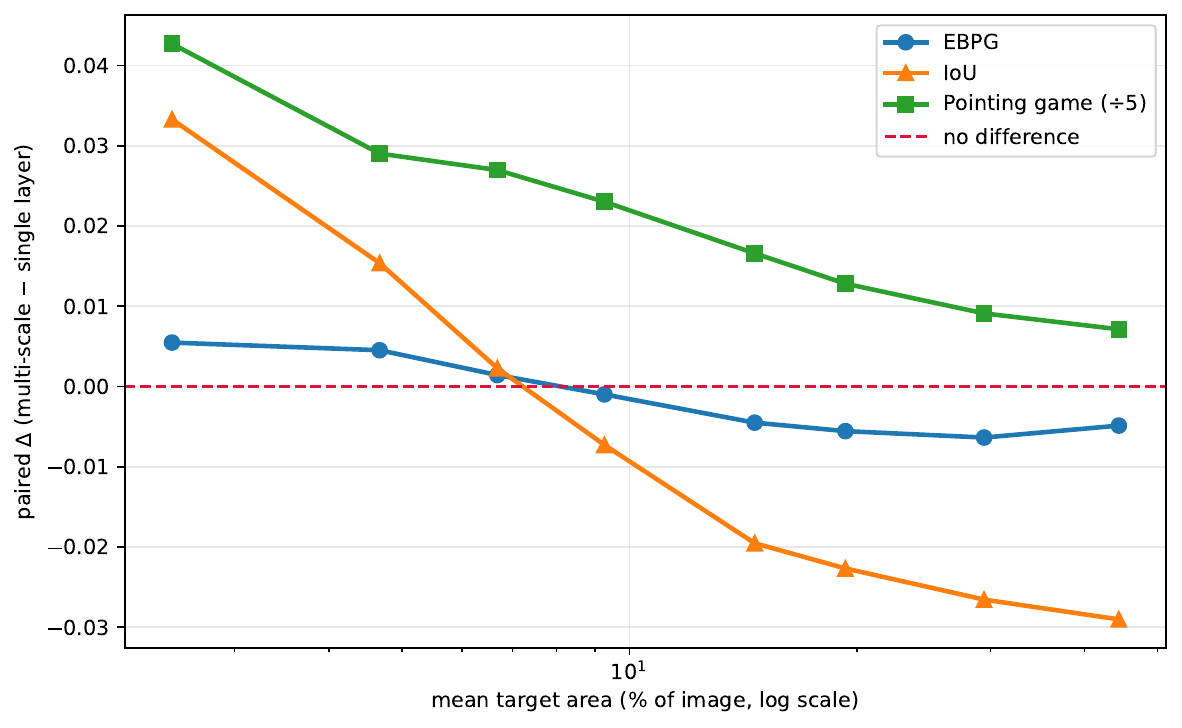}
\caption{Paired difference between the multi-scale and single-layer variants on
PASCAL VOC over nested subsets of increasingly small objects. The pointing game
(rescaled for display) stays positive throughout but shrinks as objects grow,
whereas EBPG and IoU change sign.}
\label{fig:crossover}
\end{figure}

The same analysis also identifies the point at which fusion stops helping.
Restricting the paired comparison between the two variants to progressively
smaller objects shows a change of sign (Fig.~\ref{fig:crossover}). On the full VOC set, fusion is worse on
EBPG ($-0.0049$) and IoU ($-0.0290$) while better on the pointing game
($+0.0357$). Restricted to objects below $10\%$ of the image, all three metrics
favour fusion (EBPG $+0.0045$, $p=8.5\text{e-}10$; PG $+0.1452$,
$p=1.5\text{e-}5$; IoU $+0.0154$, $p=1.3\text{e-}4$), and the margins grow further
below $5\%$ (EBPG $+0.0055$, PG $+0.2136$, IoU $+0.0333$). For VOC's
$7\times7$ final grid, the sign change between $6.7\%$ and $9.3\%$ mean area
corresponds to between $3.3$ and $4.5$ occupied cells by Eq.~\eqref{eq:cells},
in agreement with the four-cell estimate of Sec.~\ref{sec:fusion}. The same
pattern appears within BRISC itself: the fusion gain in EBPG is positive for
tumours below $5\%$ of the image ($+0.0025$, win rate $64.3\%$) and slightly
negative over the full set.

The pointing game never changes sign; it only diminishes. EBPG and IoU do change
sign, which is consistent with the mechanism: adding a high-resolution level
sharpens the peak but distributes energy over a finer structure, and when the
target is large that redistribution costs more than the sharpening gains. The
size bins are a fixed logarithmic partition of the area range; the conclusion
does not depend on them, as the binning-free rank correlation above shows.

\subsection{Faithfulness and Model Dependence}
\label{sec:sanity}

\begin{table}[htbp]
\centering
\caption{Faithfulness on BRISC 2025 and PASCAL VOC 2012 ($n=180$ each). Deletion
is better when lower, insertion when higher. Best per column in \textbf{bold},
second best \underline{underlined}. The proposed method attains the best
insertion on BRISC and is second on all four VOC columns behind HiResCAM;
per-cell significance is discussed in the text.}
\label{tab:faithfulness}
\vspace{3pt}
\resizebox{\linewidth}{!}{
\begin{tabular}{lcccccccc}
\toprule
& \multicolumn{4}{c}{\textbf{BRISC 2025}} & \multicolumn{4}{c}{\textbf{PASCAL VOC 2012}}\\
\cmidrule(lr){2-5}\cmidrule(lr){6-9}
& \multicolumn{2}{c}{\textbf{Del} $\downarrow$} & \multicolumn{2}{c}{\textbf{Ins} $\uparrow$}
& \multicolumn{2}{c}{\textbf{Del} $\downarrow$} & \multicolumn{2}{c}{\textbf{Ins} $\uparrow$}\\
\cmidrule(lr){2-3}\cmidrule(lr){4-5}\cmidrule(lr){6-7}\cmidrule(lr){8-9}
\textbf{Method} & Mean & Med & Mean & Med & Mean & Med & Mean & Med\\
\midrule
SVD-Fusion CAM (multi-scale) & 0.5104 & 0.4847 & \textbf{0.8471} & \textbf{0.9171} & \underline{0.3488} & \underline{0.3121} & \underline{0.8477} & \underline{0.8943}\\
SVD-Fusion CAM (single layer) & \underline{0.4534} & 0.4166 & \underline{0.8446} & \underline{0.9048} & 0.3534 & 0.3215 & 0.8378 & 0.8891\\
SVD-Grad-CAM~\citep{thota2024svd} & 0.4608 & 0.4175 & 0.8227 & 0.8868 & 0.3575 & 0.3492 & 0.8372 & 0.8929\\
Grad-CAM~\citep{selvaraju2017grad} & 0.4652 & \textbf{0.4092} & 0.8225 & 0.8932 & 0.3613 & 0.3519 & 0.8369 & 0.8908\\
HiResCAM~\citep{draelos2020use} & \textbf{0.4417} & \underline{0.4152} & 0.8398 & 0.8952 & \textbf{0.3412} & \textbf{0.3018} & \textbf{0.8541} & \textbf{0.9039}\\
Grad-CAM++~\citep{chattopadhay2018gradcam} & 0.5154 & 0.5149 & 0.7660 & 0.8472 & 0.3724 & 0.3666 & 0.8211 & 0.8731\\
LayerCAM~\citep{jiang2021layercam} & 0.4918 & 0.4822 & 0.8057 & 0.8891 & 0.3705 & 0.3507 & 0.8130 & 0.8715\\
PCG-CAM~\citep{huang2025pcg} & 0.5048 & 0.4853 & 0.7933 & 0.8761 & 0.3992 & 0.3840 & 0.7929 & 0.8554\\
XGrad-CAM~\citep{fu2020axiom} & 0.5494 & 0.5266 & 0.6945 & 0.7340 & 0.5507 & 0.5883 & 0.6744 & 0.7411\\
Eigen-CAM~\citep{muhammad2020eigen} & 0.5559 & 0.6054 & 0.7061 & 0.7420 & 0.5725 & 0.6358 & 0.6398 & 0.6954\\
\bottomrule
\end{tabular}}
\end{table}

As Table~\ref{tab:faithfulness} shows, the proposed method attains the highest
insertion score on BRISC and is not significantly different from the strongest baseline,
HiResCAM, on VOC. On BRISC the paired tests place it significantly above every
other baseline (all $p < 10^{-3}$) and nominally above HiResCAM ($p = 0.04$,
a margin that does not survive multiple-comparison correction); on VOC it is
significantly above every baseline except HiResCAM, from which it is
indistinguishable ($p = 0.46$). The method is therefore at least as faithful under insertion as the most
faithful gradient baseline, while retaining its localisation advantage
(Sec.~\ref{sec:sota}).

Deletion behaves differently on the two datasets, and the difference is
informative. On the medical dataset the multi-scale variant scores worse than
most baselines, which follows directly from the same mechanism that produces
the localisation gain: deletion removes pixels in order of importance, and a
finely structured map removes scattered pixels that degrade the image less
coherently than a contiguous blob. The effect is specific to the medical
domain. On PASCAL VOC, where deletion is more reliable because removing pixels
stays closer to the training distribution, the proposed method is never
significantly worse than any baseline. Replacing tissue regions produces inputs
far outside the training distribution, so the disadvantage appears only where
the metric is least trustworthy, and we regard it as a property of
medical-image perturbation rather than of the map. The single-layer variant,
which produces the coarser map, is accordingly the better choice when deletion
is the metric of interest, and Sec.~\ref{sec:practical} states when each
variant should be used.

\begin{table}[htbp]
\centering
\caption{Cascading weight randomisation on BRISC 2025 ($n=60$), mean
$\pm$ standard deviation over five random draws (seeds 0--4). Entries are
$|\rho_{\mathrm{Spearman}}|$ between the map from the trained model and the map
after randomising all layers up to and including the stated stage. We report
three representative depths, namely the classifier head, a mid-level block and
the shallowest block used by the configuration. Lower is better; the lowest mean
in each row is shown in bold.}
\label{tab:sanity}
\vspace{3pt}
\resizebox{\linewidth}{!}{
\begin{tabular}{lcccccccccc}
\toprule
\textbf{Stage} & \textbf{Ours} & \textbf{Ours} & Grad- & Grad- & HiRes & Layer & Eigen & SVD- & XGrad & PCG-\\
 & \textbf{(multi)} & \textbf{(single)} & CAM & CAM++ & CAM & CAM & CAM & Grad & CAM & CAM\\
\midrule
\texttt{head} (classifier) & 0.280{\scriptsize$\pm$0.015} & 0.307{\scriptsize$\pm$0.013} & 0.330{\scriptsize$\pm$0.013} & 0.683{\scriptsize$\pm$0.028} & 0.265{\scriptsize$\pm$0.025} & 0.920{\scriptsize$\pm$0.005} & 1.000{\scriptsize$\pm$0.000} & 0.331{\scriptsize$\pm$0.009} & 0.487{\scriptsize$\pm$0.025} & \textbf{0.244}{\scriptsize$\pm$0.023}\\
\texttt{block6} (mid) & 0.264{\scriptsize$\pm$0.074} & 0.348{\scriptsize$\pm$0.144} & 0.278{\scriptsize$\pm$0.086} & 0.355{\scriptsize$\pm$0.059} & 0.276{\scriptsize$\pm$0.071} & 0.553{\scriptsize$\pm$0.112} & 0.539{\scriptsize$\pm$0.217} & \textbf{0.260}{\scriptsize$\pm$0.091} & 0.275{\scriptsize$\pm$0.027} & 0.315{\scriptsize$\pm$0.056}\\
\texttt{block3} (shallow) & 0.362{\scriptsize$\pm$0.077} & 0.347{\scriptsize$\pm$0.109} & 0.302{\scriptsize$\pm$0.090} & 0.364{\scriptsize$\pm$0.045} & 0.304{\scriptsize$\pm$0.075} & 0.575{\scriptsize$\pm$0.081} & 0.387{\scriptsize$\pm$0.096} & 0.302{\scriptsize$\pm$0.078} & \textbf{0.285}{\scriptsize$\pm$0.086} & 0.309{\scriptsize$\pm$0.056}\\
\bottomrule
\end{tabular}}
\end{table}

As Table~\ref{tab:sanity} reports, randomising only the classifier head is
already decisive. Eigen-CAM and LayerCAM retain correlations of $1.000$ and
$0.920$ with their original maps, meaning their explanations are almost unchanged
when the network can no longer classify anything. Their maps are therefore
largely determined by image structure rather than by what the model learned. This
bears directly on Table~\ref{tab:loc_brisc}: LayerCAM is the strongest baseline on
the pointing game, but that score is achieved by a map that survives
randomisation of the classifier, which makes it weak evidence of explanatory
quality. Grad-CAM++ ($0.683$) and XGrad-CAM ($0.487$) fall between these two and
the remaining gradient-based methods.

At the head stage, where the variation across random draws is small (standard
deviation at most $0.028$), the proposed method ($0.280$) is close to the lowest
values ($0.244$ for PCG-CAM and $0.265$ for HiResCAM) and ranks second or third in
every draw. At the deeper stages the gradient-based methods fall to similar
values between $0.26$ and $0.36$, but the variation across draws (standard
deviations up to $0.14$) exceeds the differences between them. The test
therefore separates model-dependent methods from the edge-detector-like ones,
Eigen-CAM and LayerCAM, rather than ranking the model-dependent methods against
each other, and for the same reason we draw no conclusion from it about the
effect of multi-scale fusion.

\subsection{Qualitative Analysis}
\label{sec:qualitative}

\begin{figure*}[t]
\centering
\includegraphics[width=\textwidth]{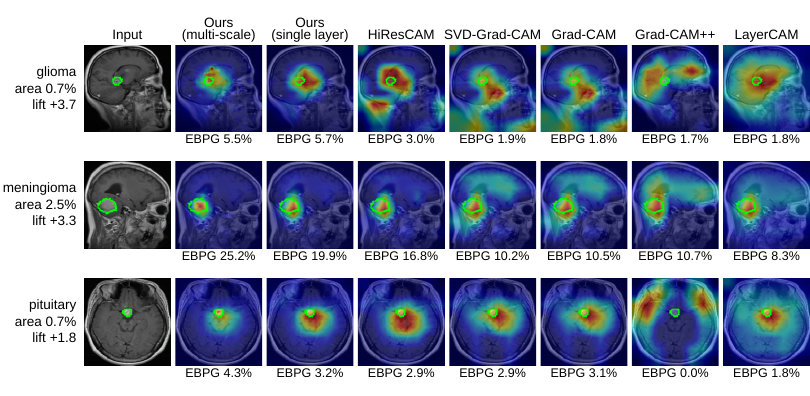}
\caption{Qualitative comparison on BRISC 2025, one case per tumour type. The
green contour marks the ground-truth boundary and the per-panel EBPG is given
below each map. Cases were selected as those where the proposed multi-scale
variant exceeds the strongest baseline by the largest margin in lift, so
that the comparison is legible; the tumour's share of the image (area) and this
margin (lift) are stated on the left of each row. Ours (multi-scale) and Ours
(single layer) are the two variants of SVD-Fusion CAM. The
single-layer variant is not part of that selection criterion and occasionally
attains a higher EBPG on these particular images, consistent with
Sec.~\ref{sec:size_law}, where fusion is shown to improve peak localisation more
than energy concentration. Aggregate performance over all 859 images is reported
in Table~\ref{tab:loc_brisc}.}
\label{fig:qual_brisc}
\end{figure*}

\begin{figure*}[t]
\centering
\includegraphics[width=\textwidth]{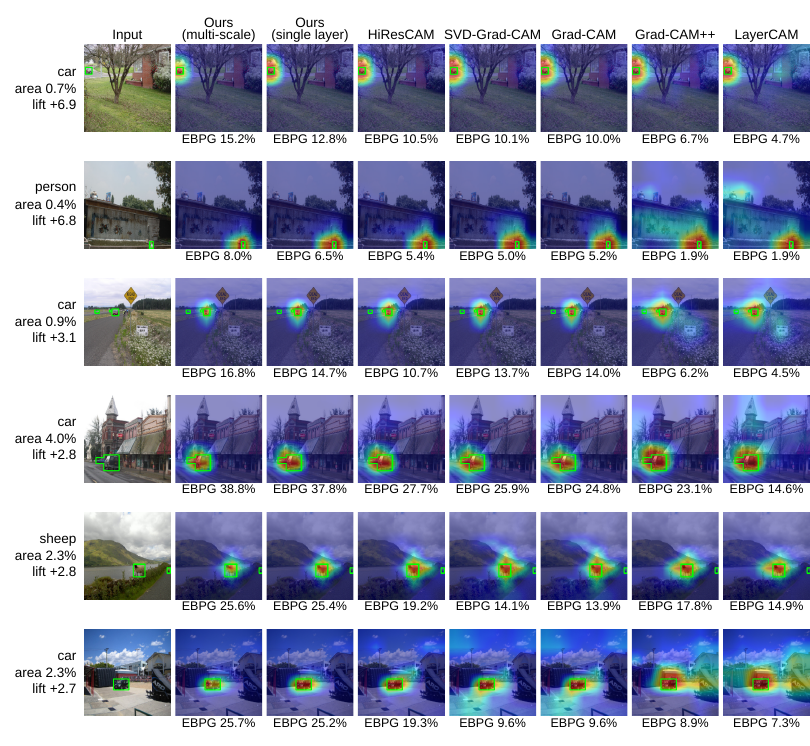}
\caption{Qualitative comparison on PASCAL VOC 2012, restricted to objects
covering less than $20\%$ of the image, the regime in which the cell-coverage
argument predicts an advantage. Cases were selected as those with the largest
margin in lift over the strongest baseline; the object's share of the image
(area) and this margin (lift) are stated on the left of each row. The green boxes
mark the ground truth, the per-panel EBPG is given below each map, and Ours
denotes the two variants of SVD-Fusion CAM. Aggregate performance over all
1401 images is reported in Table~\ref{tab:loc_voc}.}
\label{fig:qual_voc}
\end{figure*}

Figures~\ref{fig:qual_brisc} and~\ref{fig:qual_voc} show representative
comparisons. Both figures select the cases in which the proposed method exceeds
the strongest baseline by the largest margin in lift; this makes the difference
between methods legible but is a selection rule, and the margin for each row is
printed alongside it so that the reader can calibrate against the aggregate
results in Tables~\ref{tab:loc_brisc} and~\ref{tab:loc_voc}.

Two patterns are visible. On the medical images the proposed maps concentrate
inside the annotated boundary while several baselines place substantial energy on
the skull, the scalp or contralateral tissue; SVD-Grad-CAM in particular tends to
activate along the outer contour of the head. On natural images, where the
figure is restricted to objects covering less than a fifth of the frame, the
proposed maps form a compact response around the object whereas several baselines,
Grad-CAM++ and LayerCAM in particular, spread across the surrounding context, which is consistent with their higher IoU
and lower energy concentration reported in Table~\ref{tab:loc_voc}.

\subsection{Computational Cost}
\label{sec:runtime}

Table~\ref{tab:runtime} reports end-to-end times per image, including the forward
and backward passes, averaged over 30 images on a single NVIDIA T4. The table
lists the methods that differ materially in cost. HiResCAM, XGrad-CAM and
PCG-CAM are omitted because they are computationally indistinguishable from
Grad-CAM: all four require exactly one forward and one backward pass, and differ
only in a handful of operations on the resulting maps, which is negligible
against the cost of the passes themselves. For PCG-CAM this is one
eigendecomposition of a $64\times64$ covariance matrix, which takes about
$2$~ms per image even on a CPU.

\begin{table}[htbp]
\centering
\caption{Mean explanation time per image on BRISC 2025 (EfficientNet-B1,
$240\times240$ input, single NVIDIA T4). Times include the forward and backward
passes shared by all gradient-based methods.}
\label{tab:runtime}
\small
\begin{tabular}{lcc}
\toprule
\textbf{Method} & \textbf{Time (ms)} & \textbf{vs.\ Grad-CAM}\\
\midrule
LayerCAM & 916 & $1.00\times$\\
Grad-CAM & 916 & $1.00\times$\\
Grad-CAM++ & 922 & $1.01\times$\\
\textbf{SVD-Fusion CAM (multi-scale)} & 944 & $1.03\times$\\
\textbf{SVD-Fusion CAM (single layer)} & 968 & $1.06\times$\\
SVD-Grad-CAM & 1030 & $1.12\times$\\
Eigen-CAM & 4025 & $4.39\times$\\
Score-CAM & 7496 & $8.18\times$\\
\bottomrule
\end{tabular}
\end{table}

The overhead of the proposed method over Grad-CAM is $3\%$. The difference between
the two variants of the proposed method is within run-to-run variation (standard
deviation $140$--$190$~ms across images). The thin SVD operates
on a matrix whose row count is the number of feature-map cells rather than
pixels, so its cost is negligible beside the backward pass. SVD-Grad-CAM is
slower despite computing a scalar weight, because it decomposes each channel
separately rather than the joint matrix once. Our Eigen-CAM implementation
computes an eigendecomposition of the $C\times C$ activation covariance matrix,
for which a thin SVD of the $N\times C$ activation matrix would be cheaper, and
Score-CAM requires one forward pass per channel, which makes them $4.4$ and $8.2$ times slower than Grad-CAM. The method is therefore compatible with interactive use.

\FloatBarrier
\section{Practical Considerations}
\label{sec:practical}

Beyond the quantitative comparison, three aspects determine whether the method
is useful in practice: its computational cost, the guidance it offers for
choosing the explanation layer, and the role it can play within a decision
pipeline. We discuss them in turn and state which variant to use.

The first consideration is cost. The method adds one thin SVD to a single
backward pass. Measured end-to-end, it takes $944$\,ms per image against
$916$\,ms for Grad-CAM, an overhead of $3\%$, and it is roughly eight times
faster than Score-CAM, which requires one forward pass per channel. This places
it in the same regime as Grad-CAM and LayerCAM and well below perturbation-based
methods, whose cost grows with the number of channels. The overhead is therefore
compatible with interactive use, since an explanation can be produced alongside
a prediction without a separate offline stage.

A low cost is useful only if the method can also be configured without
extensive tuning, and here Eq.~\eqref{eq:cells} is directly actionable. In a
given deployment setting the expected target size relative to the input is
usually known: lesions occupy a small fraction of a scan, whereas objects in
consumer photographs often dominate the frame. Computing $n_l = \rho N_l$ for
the final layer indicates whether that layer alone can resolve the target, and
hence whether a higher-resolution level should be included. This calculation
can be performed before any explanation is generated and narrows the usual
trial-and-error choice of explanation layer; in our experiments, the
configurations found by exhaustive search on the training split agree with it
(Sec.~\ref{sec:ablation}). The rule for choosing between the two variants
follows directly: when the targets of interest occupy fewer than about four
cells of the deepest layer, the multi-scale variant should be used, since it
improves peak placement by a wide margin at a small cost in energy
concentration; when they occupy more, the single-layer variant is sufficient
and gives the more compact map, which is also the one preferred by deletion.

Once configured, the method can support decisions in two ways. In a clinical
triage setting the explanation is not the output but a check on the output.
First, an explanation whose energy falls far outside the region a clinician
would consider relevant is a signal to review the case rather than to trust the
prediction, which is useful even when the prediction is correct. Second,
aggregate localisation statistics over a validation set can reveal classifiers
that achieve high accuracy through context rather than through the target
itself, a failure mode that accuracy alone does not reveal. Sec.~\ref{sec:perclass}
discusses a possible instance of this behaviour, in which the classifier may
rely on peritumoural signal outside the annotated glioma region.

These uses also have a limit. Localisation agreement is evidence that the model
and the annotation concur; it is not evidence that the model is correct, nor
that the explanation is complete. Explanations of this family should therefore
inform review rather than replace it.

\FloatBarrier
\section{Scope and Limitations}
\label{sec:limitations}

The conclusions of this study hold within the scope of gradient-based CAM for
convolutional classifiers, and are subject to the usual limitations of
saliency evaluation. We state both.

Proposition~\ref{prop:attribution} concerns the first-order, pre-ReLU
behaviour of the explanation. It establishes that the proposed weighting is
the best rank-$k$ approximation to the gradient in the sense that matters for
faithfulness, but it does not by itself predict localisation accuracy, which
depends on where the model's evidence lies; Sec.~\ref{sec:perclass} shows a
case in which the two differ. Likewise, the cell-coverage criterion is an
upper bound on attainable precision, not a predictor of achieved precision.

The evaluation metrics have known limitations. Deletion and insertion replace
image regions with a constant or blurred substitute, producing inputs unlike
anything the model saw in training; the concern is sharper for MRI, where local
texture is diagnostic, than for natural images, which is why we weight insertion
and the localisation metrics more heavily than deletion. IoU requires
binarising a continuous map, and the ranking under this metric depends on the
threshold when targets are large relative to the image and diffuse maps are
therefore advantaged; we apply one threshold identically to all methods and
show that the ranking reverses for small targets at thresholds of $0.3$ and
$0.5$ (\ref{app:iou_thr}).

The evaluation covers two classification datasets and two EfficientNet
backbones. The method itself requires only a convolutional layer with a
spatial grid and a differentiable class score, and the cell-coverage criterion
depends only on the feature-map size, so neither is specific to this family;
extension to detection and segmentation models, where the target region is
defined differently, is left to future work.


\FloatBarrier
\section{Conclusion}
\label{sec:conclusion}

We revisited a step that gradient-based CAM methods perform almost without
comment: the reduction of each channel's two-dimensional gradient map to a single
scalar. We showed that this reduction is a projection onto a fixed spatial
direction, and that estimating the dominant direction instead, through a
singular value decomposition of the joint spatial--channel gradient matrix,
yields a weight map rather than a weight. The map minimises the first-order
attribution error among all rank-$k$ weightings, and Grad-CAM's fixed direction
can never do better than its rank-one estimate. The resulting formulation is
not disjoint from prior work: Grad-CAM is the most constrained cell of a
two-axis design space that the decomposition makes explicit, HiResCAM is its
untruncated limit, and the two variants of our method occupy the cells between.

We also argued that the choice of layers to explain is a design decision in its
own right, and gave a criterion for it. Because a map computed at a given layer
cannot resolve detail finer than one feature-map cell, the number of cells a
target occupies bounds the precision that any weighting scheme can reach. This
criterion depends only on the feature-map geometry, can be computed before any
explanation is generated, and makes a falsifiable prediction: multi-scale
fusion should help in proportion to how small the target is, with a transition
near four cells. Across two datasets whose targets differ by more than an order
of magnitude in relative size, that prediction held, and the benefit of fusion
changed sign between $3.3$ and $4.5$ cells. Against eight baselines the method
leads on localisation lift and pointing-game accuracy on both datasets, by
$14.4\%$ and $71\%$ over the strongest baseline where targets are small, with
the lift margin narrowing to $4.9\%$ where they are large, at a $3\%$ overhead
over Grad-CAM.

The analysis also states when each variant should be used. Multi-scale fusion
sharpens the peak of the explanation at the cost of a finer energy
distribution, so it is the right choice for targets that occupy few cells of
the deepest layer, while the single-layer variant is preferable for large
targets and whenever a compact map is required. We regard the ability to make
this choice from the feature-map geometry alone as more useful than a claim
that one variant is universally superior.

Future work will extend the evaluation to detection and segmentation models,
where the notion of a target region differs from classification, and will
examine whether a noise-level criterion on the gradient spectrum, such as
parallel analysis, can replace the fixed rank $k$ by an adaptive per-image
truncation.

\section*{CRediT authorship contribution statement}
\textbf{Newsha Shahbodaghkhan:} Conceptualization, Methodology, Software, Validation, Formal analysis, Investigation, Visualization, Writing - original draft.
\textbf{Maryam Amirmazlaghani:} Conceptualization, Supervision, Validation, Writing - review \& editing.
\textbf{Mohammad Rahmati:} Supervision, Writing - review \& editing.
\textbf{Mansoor Rezghi:} Conceptualization, Writing - review \& editing.

\section*{Declaration of competing interest}
The authors declare that they have no known competing financial interests or personal relationships that could have appeared to influence the work reported in this paper.

\section*{Funding}
This research did not receive any specific grant from funding agencies in the
public, commercial, or not-for-profit sectors.

\section*{Data availability}
The datasets used in this study (PASCAL VOC 2012 and BRISC 2025) are publicly available. The code and scripts that reproduce the tables and figures are available at \url{https://github.com/newshaShah/SVD_Fusion_CAM}.

\section*{Declaration of generative AI and AI-assisted technologies in the manuscript preparation process}
During the preparation of this work the authors used Claude (Anthropic) in order to
improve the language and readability of the manuscript and as a coding assistant
for parts of the evaluation code. The method, experimental design, analysis and
interpretation are the work of the authors, who reviewed and verified all
AI-assisted output and take full responsibility for the content of the
publication.

\bibliographystyle{elsarticle-harv}
\bibliography{references_merged}

@article{tjoa2020survey,
	title={A survey on explainable artificial intelligence ({XAI}): Toward medical {XAI}},
	author={Tjoa, Erico and Guan, Cuntai},
	journal={IEEE transactions on neural networks and learning systems},
	volume={32},
	number={11},
	pages={4793--4813},
	year={2021},
	publisher={IEEE},
	doi={10.1109/TNNLS.2020.3027314}
}

@inproceedings{zhou2016learning,
	title={Learning deep features for discriminative localization},
	author={Zhou, Bolei and Khosla, Aditya and Lapedriza, Agata and Oliva, Aude and Torralba, Antonio},
	booktitle={Proceedings of the IEEE conference on computer vision and pattern recognition},
	pages={2921--2929},
	year={2016},
	doi={10.1109/CVPR.2016.319}
}

@inproceedings{selvaraju2017grad,
	title={{Grad-CAM}: Visual explanations from deep networks via gradient-based localization},
	author={Selvaraju, Ramprasaath R and Cogswell, Michael and Das, Abhishek and Vedantam, Ramakrishna and Parikh, Devi and Batra, Dhruv},
	booktitle={Proceedings of the IEEE international conference on computer vision},
	pages={618--626},
	year={2017},
	doi={10.1109/ICCV.2017.74}
}

@inproceedings{chattopadhay2018gradcam,
	title={{Grad-CAM++}: Generalized gradient-based visual explanations for deep convolutional networks},
	author={Chattopadhay, Aditya and Sarkar, Anirban and Howlader, Prantik and Balasubramanian, Vineeth N},
	booktitle={2018 IEEE Winter Conference on Applications of Computer Vision (WACV)},
	pages={839--847},
	year={2018},
	organization={IEEE},
	doi={10.1109/WACV.2018.00097}
}

@inproceedings{wang2020scorecam,
	title={{Score-CAM}: Score-weighted visual explanations for convolutional neural networks},
	author={Wang, Haofan and Wang, Zifan and Du, Mengnan and Yang, Fan and Zhang, Zijian and Ding, Sirui and Mardziel, Piotr and Hu, Xia},
	booktitle={Proceedings of the IEEE/CVF conference on computer vision and pattern recognition workshops},
	pages={111--119},
	year={2020},
	doi={10.1109/CVPRW50498.2020.00020}
}

@article{everingham2010pascal,
	title={The {PASCAL} Visual Object Classes ({VOC}) challenge},
	author={Everingham, Mark and Van Gool, Luc and Williams, Christopher KI and Winn, John and Zisserman, Andrew},
	journal={International Journal of Computer Vision},
	volume={88},
	number={2},
	pages={303--338},
	year={2010},
	publisher={Springer},
	doi={10.1007/s11263-009-0275-4}
}

@inproceedings{fu2020axiom,
	title={Axiom-based {Grad-CAM}: Towards accurate visualization and explanation of {CNNs}},
	author={Fu, Ruigang and Hu, Qingyong and Dong, Xiaobo and Guo, Yulan and Gao, Yue and Li, Biao},
	booktitle={British Machine Vision Conference (BMVC)},
	year={2020}
}

@inproceedings{muhammad2020eigen,
	title={{Eigen-CAM}: Class activation map using principal components},
	author={Muhammad, Mohammed Bany and Yeasin, Mohammed},
	booktitle={2020 international joint conference on neural networks (IJCNN)},
	pages={1--7},
	year={2020},
	organization={IEEE},
	doi={10.1109/IJCNN48605.2020.9206626}
}

@article{jiang2021layercam,
	title={{LayerCAM}: Exploring hierarchical class activation maps for localization},
	author={Jiang, Peng-Tao and Zhang, Chang-Bin and Hou, Qibin and Cheng, Ming-Ming and Wei, Yunchao},
	journal={IEEE transactions on image processing},
	volume={30},
	pages={5875--5888},
	year={2021},
	publisher={IEEE},
	doi={10.1109/TIP.2021.3089943}
}

@inproceedings{ramaswamy2020ablation,
  title={{Ablation-CAM}: Visual explanations for deep convolutional network via gradient-free localization},
  author={Desai, Saurabh and Ramaswamy, Harish Guruprasad},
  booktitle={Proceedings of the IEEE/CVF Winter Conference on Applications of Computer Vision (WACV)},
  pages={983--991},
  year={2020}
}

@article{srinivas2019full,
	title={Full-gradient representation for neural network visualization},
	author={Srinivas, Suraj and Fleuret, Fran{\c{c}}ois},
	journal={Advances in neural information processing systems},
	volume={32},
	year={2019}
}

@inproceedings{thota2024svd,
  title     = {{SVD-Grad-CAM}: Singular Value Decomposition filtered Gradient Weighted Class Activation Map},
  author    = {Thota, Gokaramaiah and Nagaraju, K. and Korra, Sathya Babu},
  booktitle = {Pattern Recognition: 27th International Conference, ICPR 2024, Proceedings, Part XII},
  pages     = {90--105},
  year      = {2024},
  publisher = {Springer},
  doi       = {10.1007/978-3-031-78198-8\_7}
}

@article{arrieta2020explainable,
	title={Explainable Artificial Intelligence ({XAI}): Concepts, taxonomies, opportunities and challenges toward responsible {AI}},
	author={Arrieta, Alejandro Barredo and D{\'\i}az-Rodr{\'\i}guez, Natalia and Del Ser, Javier and Bennetot, Adrien and Tabik, Siham and Barbado, Alberto and Garc{\'\i}a, Salvador and Gil-L{\'o}pez, Sergio and Molina, Daniel and Benjamins, Richard and others},
	journal={Information fusion},
	volume={58},
	pages={82--115},
	year={2020},
	publisher={Elsevier},
	doi={10.1016/j.inffus.2019.12.012}
}

@article{adebayo2018sanity,
	title={Sanity checks for saliency maps},
	author={Adebayo, Julius and Gilmer, Justin and Muelly, Michael and Goodfellow, Ian and Hardt, Moritz and Kim, Been},
	journal={Advances in neural information processing systems},
	volume={31},
	year={2018}
}

@book{golub2013matrix,
	title={Matrix computations},
	author={Golub, Gene H and Van Loan, Charles F},
	year={2013},
	publisher={JHU press}
}

@article{zhang2018top,
	title={Top-down neural attention by excitation backprop},
	author={Zhang, Jianming and Bargal, Sarah Adel and Lin, Zhe and Brandt, Jonathan and Shen, Xiaohui and Sclaroff, Stan},
	journal={International Journal of Computer Vision},
	volume={126},
	number={10},
	pages={1084--1102},
	year={2018},
	publisher={Springer},
	doi={10.1007/s11263-017-1059-x}
}

@inproceedings{tan2019efficientnet,
	title={{EfficientNet}: Rethinking model scaling for convolutional neural networks},
	author={Tan, Mingxing and Le, Quoc},
	booktitle={International conference on machine learning},
	pages={6105--6114},
	year={2019},
	organization={PMLR}
}

@article{van2022explainable,
	title={Explainable artificial intelligence ({XAI}) in deep learning-based medical image analysis},
	author={Van der Velden, Bas HM and Kuijf, Hugo J and Gilhuijs, Kenneth GA and Viergever, Max A},
	journal={Medical image analysis},
	volume={79},
	pages={102470},
	year={2022},
	publisher={Elsevier},
	doi={10.1016/j.media.2022.102470}
}

@article{brisc2025,
  title     = {{BRISC}: Annotated dataset for brain tumor segmentation and
               classification},
  author    = {Fateh, Amirreza and Rezvani, Yasin and Moayedi, Sara and Rezvani, Sadjad and Fateh, Fatemeh and Fateh, Mansoor and Abolghasemi, Vahid},
  journal   = {Scientific Data},
  volume    = {13},
  number    = {1},
  pages     = {361},
  year      = {2026},
  publisher = {Nature Publishing Group},
  doi       = {10.1038/s41597-026-06753-y},
  note      = {Dataset available at \url{https://doi.org/10.34740/KAGGLE/DS/7632487}}
}

@inproceedings{petsiuk2018rise,
  title     = {{RISE}: Randomized Input Sampling for Explanation of Black-box
               Models},
  author    = {Petsiuk, Vitali and Das, Abir and Saenko, Kate},
  booktitle = {British Machine Vision Conference (BMVC)},
  year      = {2018}
}

@article{zhang2024opticam,
  title   = {{Opti-CAM}: Optimizing saliency maps for interpretability},
  author  = {Zhang, Hanwei and Torres, Felipe and Sicre, Ronan and Avrithis, Yannis and Ayache, Stephane},
  journal = {Computer Vision and Image Understanding},
  volume  = {248},
  pages   = {104101},
  year    = {2024},
  doi     = {10.1016/j.cviu.2024.104101}
}

@inproceedings{byun2022recipro,
  title     = {{ReciproCAM}: Lightweight Gradient-free Class Activation Map for Post-hoc Explanations},
  author    = {Byun, Seok-Yong and Lee, Wonju},
  booktitle = {Proceedings of the IEEE/CVF Conference on Computer Vision and Pattern Recognition (CVPR) Workshops},
  pages     = {8364--8370},
  year      = {2024}
}

@misc{draelos2020use,
  title={Use {HiResCAM} instead of {Grad-CAM} for faithful explanations of convolutional neural networks},
  author={Draelos, Rachel Lea and Carin, Lawrence},
  howpublished={arXiv preprint arXiv:2011.08891},
  year={2020}
}

@article{wall2026winsor,
  author  = {Wall, Casey and Wang, Longwei and Rizk, Rodrigue and Santosh, KC},
  title   = {{Winsor-CAM}: Human-Tunable Visual Explanations From Deep Networks via Layer-Wise Winsorization},
  journal = {IEEE Transactions on Pattern Analysis and Machine Intelligence},
  year    = {2026},
  volume  = {48},
  number  = {7},
  pages   = {7920--7937},
  doi     = {10.1109/TPAMI.2026.3668075}
}

@misc{bouayed2024cnn,
  title={{CNN} Explainability with Multivector {Tucker} Saliency Maps for Self-Supervised Models},
  author={Bouayed, Aymene Mohammed and Deslauriers-Gauthier, Samuel and Iaccovelli, Adrian and Naccache, David},
  howpublished={arXiv preprint arXiv:2410.23072},
  year={2024}
}

@article{huang2025pcg,
  title={{PCG-CAM}: Enhanced class activation map using principal components of gradients and its applications in brain {MRI}},
  author={Huang, Lan and Shao, Yangguang and Hou, Wenju and Yang, Hui and Wang, Yan and Sheng, Nan and Sun, Yinglu and Wang, Yao},
  journal={Information Sciences},
  volume={708},
  pages={122046},
  year={2025},
  publisher={Elsevier},
  doi={10.1016/j.ins.2025.122046}
}

@inproceedings{shi2020zoom,
  title={{Zoom-CAM}: Generating fine-grained pixel annotations from image labels},
  author={Shi, Xiangwei and Khademi, Seyran and Li, Yunqiang and van Gemert, Jan},
  booktitle={25th International Conference on Pattern Recognition (ICPR)},
  pages={10289--10296},
  year={2021},
  organization={IEEE}
}

@article{englebert2024poly,
  title={{Poly-CAM}: High resolution class activation map for convolutional neural networks},
  author={Englebert, Alexandre and Cornu, Olivier and De Vleeschouwer, Christophe},
  journal={Machine Vision and Applications},
  volume={35},
  number={4},
  pages={89},
  year={2024},
  publisher={Springer},
  doi={10.1007/s00138-024-01567-7}
}

@article{zhang2025ipcam,
	title={{IP-CAM}: Class activation mapping based on importance weights and principal-component weights for better and simpler visual explanations},
	author={Zhang, Wenyi and Zhang, Haoran and Zhang, Xisheng and Shen, Xiaohua and Zou, Lejun},
	journal={Computer Vision and Image Understanding},
	volume={261},
	pages={104523},
	year={2025},
	publisher={Elsevier},
	doi={10.1016/j.cviu.2025.104523}
}

\clearpage
\appendix
\section{Paired Comparisons on PASCAL VOC}
\label{app:paired_voc}

\begin{table}[htbp]
\centering
\caption{Paired comparison on PASCAL VOC 2012 ($n=1401$) between SVD-Fusion
CAM (multi-scale) and every other method, using the Wilcoxon signed-rank test
on identical images. Lift, the pointing game and IoU are the localisation metrics of
Sec.~\ref{sec:metrics}. Win rate is the percentage of images on which the
proposed method scores strictly higher, and the effect size $r$ is the
matched-pairs rank-biserial correlation, which ranges from $-1$ to $+1$ and is
positive when the proposed method is better. Large effects in favour of the
proposed method ($r\ge 0.5$) are shown in bold.}
\label{tab:paired_voc}
\resizebox{\linewidth}{!}{
\begin{tabular}{llccc}
\toprule
\textbf{Metric} & \textbf{Compared with} & \textbf{Win rate (\%)} & \textbf{Effect size $r$} & \textbf{$p$}\\
\midrule
\multirow{9}{*}{Lift}
 & SVD-Fusion CAM (single layer) & 38.5 & $-0.234^{\dagger}$ & 3.8e-14\\
 & HiResCAM & 82.5 & $\mathbf{+0.724}^{*}$ & 1.4e-121\\
 & SVD-Grad-CAM & 86.1 & $\mathbf{+0.834}^{*}$ & 4.3e-160\\
 & Grad-CAM & 87.1 & $\mathbf{+0.860}^{*}$ & 4.0e-170\\
 & PCG-CAM & 88.8 & $\mathbf{+0.868}^{*}$ & 1.1e-173\\
 & Grad-CAM++ & 96.7 & $\mathbf{+0.975}^{*}$ & 1.5e-217\\
 & LayerCAM & 93.1 & $\mathbf{+0.913}^{*}$ & 2.8e-191\\
 & XGrad-CAM & 81.3 & $\mathbf{+0.802}^{*}$ & 3.3e-148\\
 & Eigen-CAM & 68.6 & $\mathbf{+0.625}^{*}$ & 9.4e-91\\
\midrule
\multirow{9}{*}{Pointing game}
 & SVD-Fusion CAM (single layer) & 4.4 & $\mathbf{+0.694}^{*}$ & 3.8e-09\\
 & HiResCAM & 5.4 & $+0.456^{*}$ & 3.6e-06\\
 & SVD-Grad-CAM & 5.9 & $\mathbf{+0.673}^{*}$ & 2.6e-11\\
 & Grad-CAM & 6.6 & $\mathbf{+0.755}^{*}$ & 7.8e-15\\
 & PCG-CAM & 10.5 & $\mathbf{+0.670}^{*}$ & 5.9e-19\\
 & Grad-CAM++ & 7.8 & $\mathbf{+0.717}^{*}$ & 6.8e-16\\
 & LayerCAM & 12.2 & $\mathbf{+0.685}^{*}$ & 1.7e-22\\
 & XGrad-CAM & 34.5 & $\mathbf{+0.940}^{*}$ & 1.2e-97\\
 & Eigen-CAM & 40.5 & $\mathbf{+0.938}^{*}$ & 7.7e-114\\
\midrule
\multirow{9}{*}{IoU}
 & SVD-Fusion CAM (single layer) & 36.7 & $-0.488^{\dagger}$ & 5.3e-56\\
 & HiResCAM & 19.1 & $-0.772^{\dagger}$ & 9.3e-138\\
 & SVD-Grad-CAM & 15.1 & $-0.814^{\dagger}$ & 1.1e-152\\
 & Grad-CAM & 15.6 & $-0.807^{\dagger}$ & 4.2e-150\\
 & PCG-CAM & 29.2 & $-0.479^{\dagger}$ & 2.6e-54\\
 & Grad-CAM++ & 14.1 & $-0.830^{\dagger}$ & 2.9e-159\\
 & LayerCAM & 18.1 & $-0.742^{\dagger}$ & 1.4e-127\\
 & XGrad-CAM & 73.4 & $\mathbf{+0.637}^{*}$ & 1.3e-94\\
 & Eigen-CAM & 64.8 & $+0.466^{*}$ & 3.3e-51\\
\bottomrule
\end{tabular}}
\begin{minipage}{\linewidth}\vspace{2pt}\footnotesize
$^{*}$ proposed method significantly better; $^{\dagger}$ significantly worse
($p<0.05$). The IoU column reflects the large mean object area of this dataset,
under which diffuse maps are advantaged at a fixed threshold; the ordering
reverses for small objects (Sec.~\ref{sec:sota}).
\end{minipage}
\end{table}

\FloatBarrier

\section{Localisation Lift by Target-Size Bin}
\label{app:per_size}

\begin{figure}[htbp]
\centering
\includegraphics[width=\linewidth]{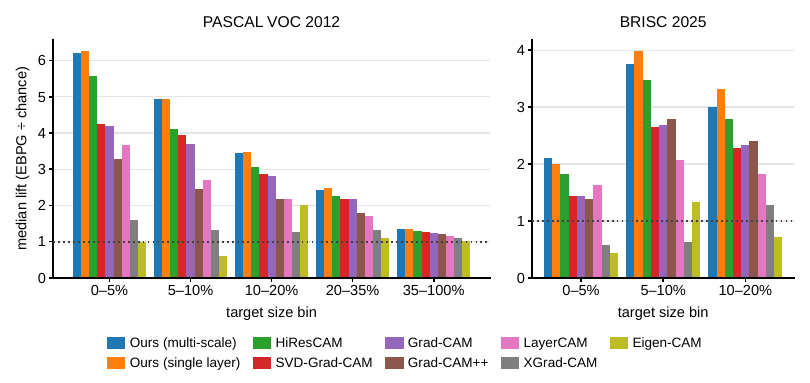}
\caption{Median localisation lift by disjoint target-size bin for both datasets.
A value of one corresponds to chance. One of the two variants of the proposed
method attains the highest lift in every bin on both datasets; the multi-scale
variant leads in the smallest BRISC bin and in the $5$--$10\%$ VOC bin, while the
single-layer variant leads elsewhere, consistent with the size analysis of
Sec.~\ref{sec:size_law}. The margin over the strongest baseline varies between
bins rather than increasing monotonically as targets shrink. PCG-CAM is not
shown; its median lift is below that of HiResCAM in every bin of both datasets,
so the strongest baseline in each bin is unchanged.}
\label{fig:per_size}
\end{figure}

Fig.~\ref{fig:per_size} complements the size analysis of
Sec.~\ref{sec:size_law} by showing the absolute localisation lift of the
compared methods in each disjoint size bin, rather than only the gain of the proposed
method over the strongest baseline.

\FloatBarrier
\section{IoU at Alternative Binarisation Thresholds}
\label{app:iou_thr}

IoU requires binarising a continuous map, and Sec.~\ref{sec:sota} attributes
the low IoU of the proposed method on the full VOC set to the large mean
object size under a fixed threshold. Table~\ref{tab:iou_thr} repeats the
measurement at thresholds of $0.3$, $0.5$ and $0.7$ of the map maximum, on the
full splits and on the VOC objects covering less than $10\%$ of the image.
Three observations follow. On BRISC the multi-scale variant is the best method
at $0.3$ and $0.5$. On the full VOC set the diffuse maps lead at every
threshold, as expected when ground truth covers $44\%$ of the image. For small
VOC objects the proposed method is the best at $0.3$ and $0.5$, and the
ordering changes only at $0.7$, where the thresholded region shrinks to the
immediate neighbourhood of the peak; the compact fused map then covers less of
the box than the broader maps of LayerCAM and HiResCAM, while the single-layer
variant stays within $2.3$ points of HiResCAM. The threshold therefore changes
the absolute values substantially but not the conclusion of
Sec.~\ref{sec:sota}: the IoU deficit of the proposed method is a property of
large targets under a fixed threshold, not of the maps.

\begin{table}[htbp]
\centering
\caption{Mean IoU (\%) at three binarisation thresholds. VOC (small) is the
subset of $186$ validation images whose annotated object covers less than
$10\%$ of the image. Best value per column in bold.}
\label{tab:iou_thr}
\resizebox{\linewidth}{!}{
\begin{tabular}{lccccccccc}
\toprule
& \multicolumn{3}{c}{\textbf{BRISC 2025} ($n=859$)} & \multicolumn{3}{c}{\textbf{VOC 2012} ($n=1401$)} & \multicolumn{3}{c}{\textbf{VOC 2012, small} ($n=186$)}\\
\cmidrule(lr){2-4}\cmidrule(lr){5-7}\cmidrule(lr){8-10}
\textbf{Method} & 0.3 & 0.5 & 0.7 & 0.3 & 0.5 & 0.7 & 0.3 & 0.5 & 0.7\\
\midrule
\textbf{SVD-Fusion CAM (multi-scale)} & \textbf{16.3} & \textbf{15.4} & 7.3 & 42.9 & 26.1 & 10.6 & \textbf{29.5} & \textbf{33.3} & 20.8\\
\textbf{SVD-Fusion CAM (single layer)} & 13.6 & 14.9 & 11.3 & 44.1 & 29.0 & 14.5 & 27.0 & 31.7 & 25.6\\
HiResCAM & 12.2 & 14.4 & 11.1 & 50.2 & 34.8 & 17.4 & 26.8 & 32.6 & 27.9\\
Grad-CAM++ & 8.9 & 12.8 & 11.9 & \textbf{52.1} & \textbf{42.8} & \textbf{23.6} & 15.9 & 25.4 & 28.9\\
LayerCAM & 8.0 & 14.7 & \textbf{15.9} & 51.9 & 39.8 & 20.8 & 19.8 & 31.1 & \textbf{32.7}\\
Grad-CAM & 9.0 & 11.8 & 9.9 & 50.3 & 37.7 & 20.1 & 22.2 & 28.5 & 26.2\\
PCG-CAM & 8.9 & 12.8 & 11.0 & 43.4 & 31.4 & 16.7 & 23.1 & 31.2 & 28.0\\
SVD-Grad-CAM & 8.9 & 11.7 & 9.9 & 50.3 & 37.5 & 19.9 & 22.7 & 29.0 & 26.4\\
XGrad-CAM & 4.1 & 4.2 & 3.3 & 23.8 & 15.0 & 7.3 & 10.6 & 12.3 & 10.0\\
Eigen-CAM & 5.1 & 6.2 & 6.4 & 24.0 & 17.1 & 9.5 & 13.2 & 16.2 & 14.5\\
\bottomrule
\end{tabular}}
\end{table}

\FloatBarrier
\section{Retained Gradient Energy and the Choice of Rank}
\label{app:energy}

Proposition~\ref{prop:attribution} bounds the first-order attribution error of
the rank-$k$ weighting by the discarded spectral energy $1-\eta_k$.
Table~\ref{tab:energy} reports $\eta_k$ of the gradient matrix $\mathbf{G}$ on
every evaluated image, for the two layers used by each configuration. The
spectrum decays slowly: at the deepest layer the leading mode retains $30\%$
of the energy on BRISC and $35\%$ on VOC, and five modes retain $63\%$ and
$69\%$; at the mid-resolution layer, where $N$ is larger, the leading mode
retains $14\%$ on both datasets. Two consequences follow. First, the
weighting used in the paper discards most of the gradient energy, so its
localisation advantage over HiResCAM, which discards none, is not a
denoising effect but a concentration effect: the dominant modes carry the
spatially coherent part of the gradient. Second, no energy threshold can
serve as a selector for $k$, since any threshold above $0.7$ selects the
largest rank considered. Table~\ref{tab:energy} therefore also reports the
proposed method with $k=5$ on the evaluation split: lift changes by at most
$3\%$ relative to the selected configuration and the pointing game by less
than one point, confirming that $k$ is a low-sensitivity parameter over the
range $\{1,\dots,5\}$.

\begin{table}[htbp]
\centering
\caption{Retained gradient energy $\eta_k$ (median over images, interquartile
range in brackets) at the two layers of each configuration, and the
localisation scores of the proposed multi-scale variant when $k=5$ is used at
both layers instead of the selected rank. The selected configurations are
$k=3$ (BRISC) and $k=1$ (VOC); their scores are those of
Tables~\ref{tab:loc_brisc} and~\ref{tab:loc_voc}.}
\label{tab:energy}
\resizebox{\linewidth}{!}{
\begin{tabular}{llccc}
\toprule
\textbf{Dataset} & \textbf{Layer} & $\eta_1$ & $\eta_3$ & $\eta_5$\\
\midrule
BRISC 2025 & $30\times30$ & 0.135 [0.122--0.154] & 0.319 [0.301--0.347] & 0.448 [0.430--0.475]\\
BRISC 2025 & $8\times8$ & 0.300 [0.281--0.321] & 0.511 [0.492--0.534] & 0.626 [0.606--0.649]\\
PASCAL VOC 2012 & $28\times28$ & 0.135 [0.118--0.167] & 0.321 [0.293--0.365] & 0.453 [0.421--0.497]\\
PASCAL VOC 2012 & $7\times7$ & 0.349 [0.324--0.382] & 0.572 [0.539--0.609] & 0.686 [0.659--0.721]\\
\midrule
\textbf{Dataset} & \textbf{Rank} & Lift (median) & PG (\%) & IoU (\%)\\
\midrule
BRISC 2025 & selected ($k=3$) & 2.31 & 43.31 & 15.42\\
BRISC 2025 & $k=5$ & 2.31 & 44.24 & 15.84\\
PASCAL VOC 2012 & selected ($k=1$) & 1.71 & 94.72 & 26.10\\
PASCAL VOC 2012 & $k=5$ & 1.66 & 94.86 & 27.55\\
\bottomrule
\end{tabular}}
\end{table}

\end{document}